\documentclass[11pt,a4paper]{article}
\usepackage[hyperref]{naaclhlt2019}
\usepackage{times}
\usepackage{latexsym}
\usepackage{subfiles}
\usepackage{amsmath}
\usepackage{amsfonts}
\usepackage{url}
\usepackage{cleveref}
\usepackage{natbib}
\usepackage{microtype}
\usepackage{xcolor}
\usepackage{booktabs}
\usepackage{tabularx}
\usepackage{multirow}
\usepackage{arydshln}
\usepackage{svg}

\usepackage{graphicx}
\graphicspath{{imgs/}{../imgs/}}

\aclfinalcopy 

\title{A Systematic Study of Leveraging Subword Information \\ for Learning Word Representations}
\author{Yi Zhu, Ivan Vuli\'c, {\normalfont and} Anna Korhonen\\
Language Technology Lab\\ University of Cambridge\\
{\tt \{yz568, iv250, alk23\}@cam.ac.uk}}
  
\date{}

\begin{document}
\maketitle

\begin{abstract}
The use of subword-level information (e.g., characters, character n-grams, morphemes) has become ubiquitous in modern word representation learning. Its importance is attested especially for morphologically rich languages which generate a large number of rare words. Despite a steadily increasing interest in such \textit{subword-informed word representations}, their systematic comparative analysis across typologically diverse languages and different tasks is still missing. In this work, we deliver such a study focusing on the variation of two crucial components required for subword-level integration into word representation models: 1) \textit{segmentation of words} into subword units, and 2) \textit{subword composition functions} to obtain final word representations. We propose a general framework for learning subword-informed word representations that allows for easy experimentation with different segmentation and composition components, also including more advanced techniques based on position embeddings and self-attention. Using the unified framework, we run experiments over a large number of subword-informed word representation configurations (60 in total) on 3 tasks (general and rare word similarity, dependency parsing, fine-grained entity typing) for 5 languages representing 3 language types. Our main results clearly indicate that there is no ``one-size-fits-all'' configuration, as performance is both language- and task-dependent. We also show that configurations based on unsupervised segmentation (e.g., BPE, Morfessor) are sometimes comparable to or even outperform the ones based on supervised word segmentation.
\end{abstract}



\section{Introduction}
Word representations are central to a wide variety of NLP tasks \cite[\textit{inter alia}]{Collobert:2011:NLP:1953048.2078186, D14-1082,P16-1002,Ammar2016ManyLO,Goldberg:2017mc,Peters:2018,P18-1007}. 
Standard word representation models are based on the {distributional hypothesis} \citep{harris54} and induce representations from large unlabeled corpora using word co-occurrence statistics \citep{DBLP:conf/nips/MikolovSCCD13,pennington2014glove,Levy:2014:NWE:2969033.2969070}. However, as pointed out by recent work \cite{Q17-1010,P17-1184,Pinter:2017emnlp,Chaudhary:2018emnlp,Zhao:2018emnlp}, mapping a finite set of word types into corresponding word representations limits the capacity of these models to learn beyond distributional information, which leads to several fundamental limitations. 

The standard approaches ignore the internal structure of words, that is, the syntactic or semantic composition from subwords or morphemes to words, and are incapable of parameter sharing at the level of subword units. Assigning only a single vector to each word causes the data sparsity problem, especially in resource-poor settings where huge amounts of training data cannot be guaranteed. The issue is also prominent for morphologically rich languages (e.g., Finnish) with productive morphological systems that generate a large number of infrequent/rare words \cite{Gerz:2018emnlp}. Although potentially useful information on word relationships is hidden in their internal \textit{subword-level structure},\footnote{For example, nouns in Finnish have 15 cases and 3 plural forms; Spanish verbs may contain over 40 inflected forms, sharing the lemma and taking up standard suffixes.} subword-agnostic word representation models do not take these structure features into account and are effectively unable to represent rare words accurately, or unseen words at all. 


\begin{figure}[t]
	\centering
	\includegraphics[width=0.99\columnwidth, trim={0.5cm 0.5cm 0 0.5cm}, clip]{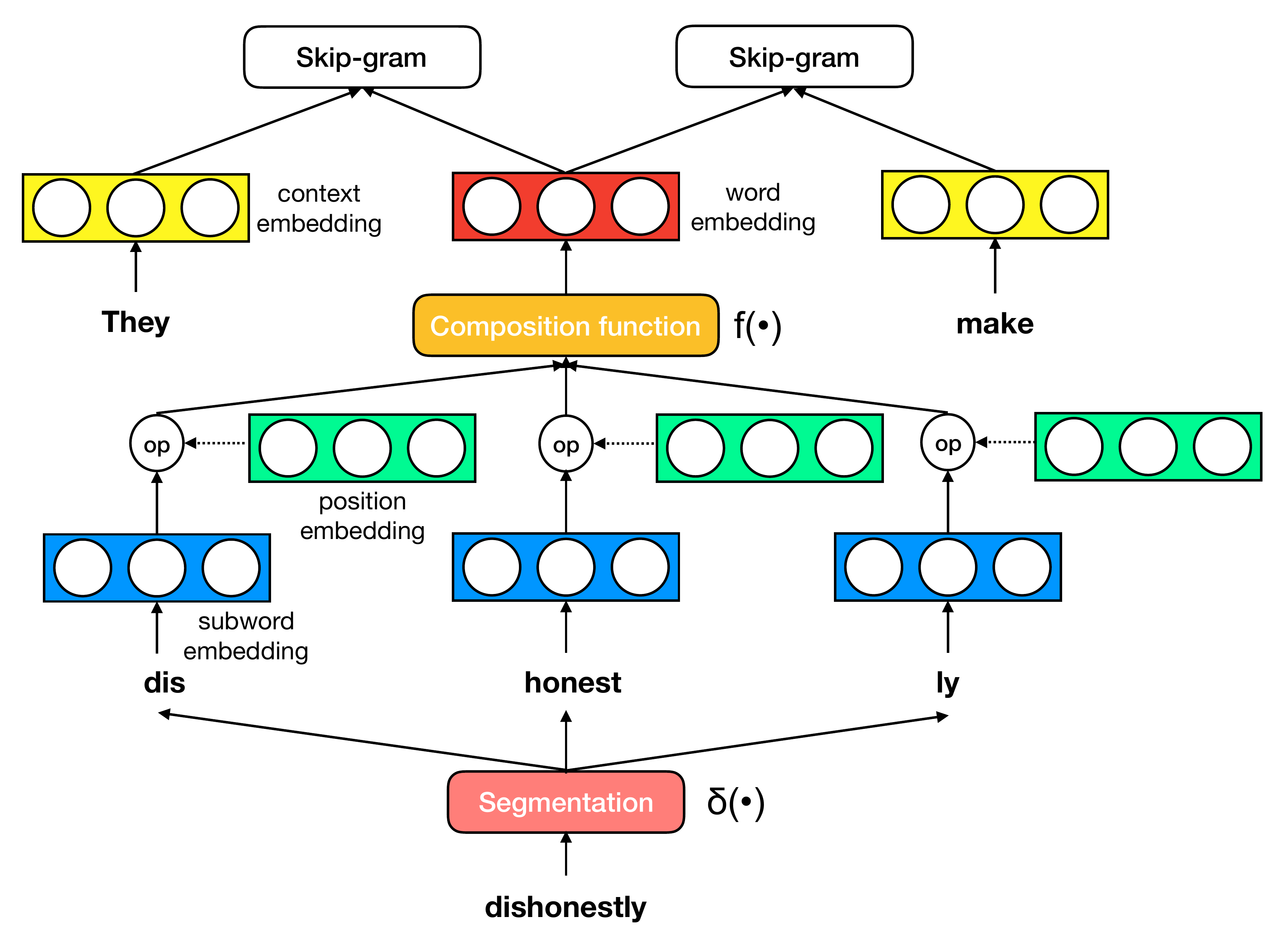}
	\caption{Illustration of the general framework for learning subword-informed word representations, with the focus on two crucial components: 1) segmentation of words and 2) subword embedding composition. By varying the two components, and optionally including or excluding position embeddings from the computations, we obtain a wide spectrum of different subword-informed configurations used in the study (see \S\ref{sec:method}). Our word-level representation model in this work is skip-gram (on the top layer of the figure), but it can be replaced by any other distributional word-level model. }\label{fig:model_arc}
\end{figure}

Therefore, there has been a surge of interest in \textit{subword-informed} word representation architectures aiming to address these gaps. A large number of architectures has been proposed in related research, and they can be clustered over the two main axes \cite{DBLP:conf/acl/LazaridouMZB13,W13-3512,C14-1015,DBLP:conf/naacl/CotterellS15,DBLP:conf/emnlp/WietingBGL16,E17-2067,P17-1184,Pinter:2017emnlp,DBLP:journals/tacl/CotterellS18}. First, the models differ in the chosen method for \textit{segmenting words into subwords}. The methods range from fully supervised approaches \citep{DBLP:conf/naacl/CotterellS15} to e.g. unsupervised approaches based on BPE \citep{Heinzerling2018BPEmbTP}. Second, another crucial aspect is the \textit{subword composition function} used to obtain word embeddings from the embeddings of each word's constituent subword units. Despite a steadily increasing interest in such subword-informed word representations, their systematic comparative analysis across the two main axes, as well as across typologically diverse languages and different tasks is still missing.\footnote{A preliminary study of \newcite{P17-1184} limits its focus on the use of subwords in the language modeling task.}

In this work, we conduct a systematic study of a variety of {subword-informed} word representation architectures that all can be described by a general framework illustrated by Figure~\ref{fig:model_arc}. The framework enables straightforward experimentation with prominent word segmentation methods (e.g., BPE, Morfessor, supervised segmentation systems) as well as subword composition functions (e.g., addition, self-attention), resulting in a large number of different \textit{subword-informed configurations}.\footnote{Following a similar work on subword-agnostic word embedding learning \cite{Levy:2015tacl}, our system
design choices resulting in different configurations can be seen as a set of hyper-parameters that also have to be carefully tuned for each language and each application task.}

Our study aims at providing answers to the following crucial questions: \textbf{Q1)} How generalizable are subword-informed models across typologically diverse languages and across different downstream tasks? Do different languages and tasks require different configurations to reach peak performances or is there a single best-performing configuration? \textbf{Q2)} How important is it to choose an appropriate segmentation and composition method? How effective are more generally applicable unsupervised segmentation methods? Is it always better to resort to a supervised method, if available? \textbf{Q3)} Is there a difference in performance with and without the full word representation? Can more advanced techniques based on position embeddings and self-attention yield better task performance?

We evaluate subword-informed word representation configurations originating from the general framework in three different tasks using standard benchmarks and evaluation protocols: 1) general and rare word similarity and relatedness, 2) dependency parsing, and 3) fine-grained entity typing for 5 languages representing 3 language families (fusional, introflexive, agglutinative). We show that different tasks and languages indeed require diverse subword-informed configurations to reach peak performance: this calls for a more careful language- and task-dependent tuning of configuration components. We also show that more sophisticated configurations are particularly useful for representing rare words, and that unsupervised segmentation methods can be competitive to supervised segmentation in tasks such as parsing or fine-grained entity typing. We hope that this paper will provide useful points of comparison and comprehensive guidance for developing next-generation subword-informed word representation models for typologically diverse languages.



\section{Methodology}\label{sec:method}
The general framework for learning subword-informed word representations, illustrated by Figure~\ref{fig:model_arc}, is introduced in \S\ref{sec:model_arc}. We then describe its main components: segmentation of words into subword units (\S\ref{sec:segmentations}), subword and position embeddings (\S\ref{sec:sp_emb}), and subword embedding composition functions (\S\ref{sec:comp}), along with all the configurations for these components used in our evaluation. 

\subsection{General Framework} \label{sec:model_arc}

Formally, given a word $w$, its word embedding $\mathbf{w}$ can be computed by the composition of its subword embeddings as follows:
\begin{align}
\mathbf{w} = f_{\mathbf{\Theta}}(\delta(w), \mathbf{W}_s, \mathbf{W}_p),
\label{eq:gm}
\end{align}
\noindent where $\delta(w)$ is a deterministic function that segments $w$ into an ordered sequence of its constituent subword units $S_w=(s_{w_i})_1^n$, with $s_{w_i} \in \mathcal{S}$ being a subword type from the subword vocabulary $\mathcal{S}$ of size $|\mathcal{S}|$.
Optionally, some segmentation methods can also generate a sequence of the corresponding morphotactic tags $T_w=(t_{w_i})_1^n$. 
Alone or together with $T_w$, $S_w$ is embedded into a sequence of subword representations $\mathbf{S}_w=(\mathbf{s}_{w_i})_1^n$ from the subword embedding matrix $\mathbf{W}_s \in \mathbb{R}^{|\mathcal{S}| \times d}$, where $d$ is the dimensionality of subword embeddings.
Another optional step is to obtain a sequence of position embeddings $\mathbf{P}_w = (\mathbf{p}_{w_i})_1^n$: they are taken from the position embedding matrix $\mathbf{W}_p \in \mathbb{R}^{p \times d}$, where $p$ is the maximum number of the unique positions. $\mathbf{P}_w$ can interact with $\mathbf{S}_w$ to compute the final representations for subwords $\mathbf{R}_w = (\mathbf{r}_{w_i})_1^n$ \cite{NIPS2017_7181}.
$f_{\mathbf{\Theta}}$ is a composition function taking $\mathbf{R}_w$ as input and outputting a single vector $\mathbf{w}$ as the word embedding of $w$.

For the distributional ``word-level'' training, similar to prior work \cite{Q17-1010}, we adopt the standard skip-gram with negative sampling (\textsc{sgns}) \cite{DBLP:conf/nips/MikolovSCCD13} with bag-of-words contexts. However, we note that other distributional models can also be used under the same framework. Again, following \newcite{Q17-1010}, we calculate the word embedding $\mathbf{w}_t \in \mathbb{R}^{d}$ for each target word $w_t$ using the formulation from Eq.~\eqref{eq:gm}, and parametrize context words with another word embedding matrix $\mathbf{W}_c \in \mathbb{R}^{|\mathcal{V}| \times d}$, where $|\mathcal{V}|$ is the size of word vocabulary $\mathcal{V}$.




\subsection{Segmentation of Words into Subwords} \label{sec:segmentations}
We consider three well-known segmentation methods for the function $\delta$, briefly outlined here.



\paragraph{Supervised Morphological Segmentation} 
\label{sec:seg_chipmunk}
We use CHIPMUNK \cite{DBLP:conf/conll/Cotterell0FS15} as a representative supervised segmentation system, proven to provide a good trade-off between accuracy and speed.\footnote{\url{http://cistern.cis.lmu.de/chipmunk}} It is based on semi-Markov conditional random fields \citep{NIPS2004_2648}. For each word, apart from generating $S_w$, it also outputs the corresponding morphotactic tags $T_w$.\footnote{In our experiments, we use only basic information on affixes such as prefixes and suffixes, and leave the integration of fine-grained information such as inflectional and derivational affixes as future work.}
In \S\ref{sec:sp_emb} we discuss how to incorporate information from $T_w$ into subword representations.

\paragraph{Morfessor}
Morfessor \cite{Aaltodoc:http://urn.fi/URN:NBN:fi:aalto-201409292677} denotes a family of generative probabilistic models for unsupervised morphological segmentation used, among other applications, to learn morphologically-aware word embeddings \citep{W13-3512}. 

\begin{table}[t]
	\centering
	\def\arraystretch{0.97}
    {\footnotesize
	\begin{tabularx}{\columnwidth}{l X}   
    $\delta$ & $\delta$({\it dishonestly})\\ 
	\toprule
    CHIPMUNK & ({\it dis}, {\it honest}, {\it ly})\\
    		 & ({\it prefix}, {\it root}, {\it suffix})\\
    \cmidrule(lr){1-1} \cmidrule(lr){2-2}
    Morfessor & ({\it dishonest}, {\it ly})\\
    \cmidrule(lr){1-1} \cmidrule(lr){2-2}
    BPE & ({\it dish}, {\it on}, {\it est}, {\it ly})\\
    \bottomrule
	\end{tabularx}}%
\vspace{-1mm}
\caption{Segmentations of the word {\it dishonestly}.} 
\label{tb:seg_exp} 
\end{table}

\paragraph{BPE}
Byte Pair Encoding (BPE; \citet{Gage:1994}) is a simple data compression algorithm.
It has become a de facto standard for providing subword information in neural machine translation \citep{DBLP:conf/acl/SennrichHB16a}.
The input word is initially split into a sequence of characters, with each unique character denoted as a byte.
BPE then iteratively replaces the most common pair of consecutive bytes with a new byte that does not occur within the data, and the number of iterations can be set in advance to control the granularity of the byte combinations.

An example output for all three methods is shown in Table~\ref{tb:seg_exp}. Note that a standard practice in subword-informed models is to also insert the entire word token into $S_w$ \cite{Q17-1010}.\footnote{We only do the insertion if $|S_w|>1$. For CHIPMUNK, a generic tag \texttt{word} is added to the sequence $T_w$.} This is, however, again an optional step and we evaluate configurations with and without the inclusion of the word token in $S_w$. 

\subsection{Subword and Position Embeddings} \label{sec:sp_emb}
The next step is to encode $S_w$ (or the tuple ($S_w$, $T_w$) for CHIPMUNK) to construct a sequence of subword representations $\mathbf{S}_w$. Each row of the subword embedding matrix $\mathbf{W}_s$ is simply defined as the embedding of a unique subword. For CHIPMUNK, we define each row in $\mathbf{W}_s$ as the concatenation of the subword $s$ and its predicted tag $t$. We also test CHIPMUNK configurations without the use of $T_w$ to analyze its contribution.\footnote{The extra information on tags should lead to a more expressive model resolving subword ambiguities. For instance, the subword {\it post} in {\it postwar} and noun {\it post} are intrinsically different: the former is the prefix and the later is the root.} 

After generating $\mathbf{S}_w$, an optional step is to have a \textit{learnable} position embedding sequence $\mathbf{P}_w$ further operate on $\mathbf{S}_w$ to encode the order information. Similar to $\mathbf{W}_s$, the definition of the position embedding matrix $\mathbf{W}_p$ also varies: for Morfessor and BPE, we use the absolute positions of subwords in the sequence $S_w$, whereas for CHIPMUNK morphotactic tags are encoded directly as positions.

Finally, following prior work \citep{Gehring2017ConvolutionalST,DBLP:conf/lrec/MikolovGBPJ18}, we use addition and element-wise multiplication between each subword vector $\mathbf{s}$ from $\mathbf{S}_w$ and the corresponding position vector $\mathbf{p}$ from $\mathbf{P}_w$ to compute each entry $\mathbf{r}$ for the final sequence of subword vectors $\mathbf{R}_w$:
\begin{align}
\mathbf{r} = \mathbf{s} + \mathbf{p} \hspace{4mm} \textnormal{or} \hspace{4mm} \mathbf{r} = \mathbf{s} \odot \mathbf{p}. 
\end{align}

\subsection{Composition Functions} \label{sec:comp}
A composition function $f_\mathbf{\Theta}$ is then applied to the sequence of subword embeddings $\mathbf{R}_w$ to compute the final word embedding $\mathbf{w}$.
We investigate three composition functions: 1) addition, 2) single-head and 3) multi-head self-attention \citep{NIPS2017_7181,lin+al-2017-embed-iclr}.\footnote{Using more complex compositions based on CNNs and RNNs is also possible, but we have not observed improvement in our evaluation tasks with such compositions, which is also in line with findings from recent work \cite{Li:2018ws}.}
Addition is used in the original \texttt{fastText} model of \citet{Q17-1010}, and remains a strong baseline for many tasks. However, addition treats each subword with the same importance, ignoring semantic composition and interactions among the word's constituent subwords. 
Therefore, we propose to use a self-attention mechanism, that is, a learnable weighted addition as the composition function on subword sequences. To the best of our knowledge, we are the first to apply a self-attention mechanism to the problem of subword composition.

\paragraph{Composition Based on Self-Attention}
Our self-attention mechanism is inspired by \citet{lin+al-2017-embed-iclr}.
It is essentially a multilayer feed-forward neural network without bias term, which generates a weight matrix for the variable length input $\mathbf{R}_w$: 
\begin{align}
& \mathbf{H}_w = tanh(\mathbf{W}_{h_1}\mathbf{R}_w^T) \\
& \mathbf{A}_w = softmax(\mathbf{W}_{h_2}\mathbf{H}_w)
\end{align}
\noindent Each row of $\mathbf{A}_w$ is a weight vector for rows of $\mathbf{R}_w$, which models different aspects of semantic compositions and interactions. For the single-head self-attention, $\mathbf{A}_{w}$ degenerates to a row vector as the final attention vector $\mathbf{a}_w$. For the multi-head self-attention, we average the rows of $\mathbf{A}_w$ to generate $\mathbf{a}_w$.\footnote{We have also experimented with adding an extra transformation layer over attention matrix to generate the attention vector, but without any performance gains.}
Finally, $\mathbf{w}$ is computed as the weighted addition of subword embeddings from $\mathbf{R}_w$: $\mathbf{w} = \sum_{w_1}^{w_n} a_{w_i}\mathbf{r}_{w_i}$.



\section{Experimental Setup}\label{sec:exp}
\begin{table}[t]
	\centering
    \def\arraystretch{0.95}
    {\footnotesize
	\begin{tabularx}{\columnwidth}{r X r}
		Component & Option & Label\\
		\toprule
		Segmentation & {\textsc CHIPMUNK} & {\it sms}\\
		& Morfessor & {\it morf}\\
		& BPE & {\it bpe}\\
        \cmidrule(lr){2-3}
		
		Morphotactic tag & concatenated with & {\it st} \\
		& subword (only \textit{sms}) & \\
        \cmidrule(lr){2-3}
		
		Word token & exclusion & {\it w-}\\
		& inclusion & {\it w+}\\
        \cmidrule(lr){2-3}
		
		Position embedding & exclusion & {\it p-}\\
		& additive & {\it pp}\\
		& multiplicative & {\it mp}\\
        \cmidrule(lr){2-3}
		
		Composition function & addition & {\it add}\\
		& single head attention & {\it att}\\
		& multi-head attention & {\it mtx}\\
        \bottomrule
	\end{tabularx}}%
	\vspace{-1mm}
	\caption{Different components used to construct subword-informed configurations, and their labels.}\label{tb:config}
    \vspace{-1mm}
\end{table}

We train different subword-informed model configurations on 5 languages representing 3 morphological language types: English (\textsc{en}), German (\textsc{de}), Finnish (\textsc{fi}), Turkish (\textsc{tr}) and Hebrew (\textsc{he}), see Table~\ref{tb:wikist}. 
We then evaluate the resulting subword-informed word embeddings in three distinct tasks: 1) general and rare word similarity and relatedness, 2) syntactic parsing, and 3) fine-grained entity typing. The three tasks have been selected in particular as they require different degrees of syntactic and semantic information to be stored in the input word embeddings, ranging from a purely semantic task (word similarity) over a hybrid syntactic-semantic task of entity typing to syntactic parsing.

\paragraph{Subword-Informed Configurations}
We train a large number of subword-informed configurations by varying the segmentation method $\delta$ (\S\ref{sec:segmentations}), subword embeddings $\mathbf{W}_s$, the inclusion of position embeddings $\mathbf{W}_p$ and the operations on $\mathbf{W}_s$ (\S\ref{sec:sp_emb}), and the composition functions $f_{\mathbf{\Theta}}$ (\S\ref{sec:comp}).
The configurations are based on the following variations of constituent components: (1) For the segmentation $\delta$, we test a supervised morphological system CHIPMUNK (\textit{sms}), Morfessor ({\it morf}) and BPE ({\it bpe}). A word token can be optionally inserted into the subword sequence $S_w$ for all three segmentation methods (\textit{ww}) or left out (\textit{w-}); (2) We can only embed the subword $s$ for \textit{morf} and \textit{bpe}, while with \textit{sms} we can optionally embed the concatenation of the subword and its morphotactic tag $s:t$ (\textit{st});\footnote{Once {\it st} is applied, we do not use position embeddings anymore, because the morphotactic tags are already encoded in subword embeddings, i.e., {\it st} and {\it pp} are mutually exclusive.} (3) We test subword embedding learning without position embeddings (\textit{p-}), or we integrate them using addition ({\it pp}) or element-wise multiplication ({\it mp}); (4) For the composition function function $f_\mathbf{\Theta}$, we experiment with addition (\textit{add}), single head self-attention (\textit{att}), and multi-head self-attention ({\it mtx}).
Table~\ref{tb:config} provides an overview of all components used to construct a variety of subword-informed configurations used in our evaluation.

The variations of components from Table~\ref{tb:config} yield 24 different configurations in total for \textit{sms}, and 18 for \textit{morf} and \textit{bpe}. We use pretrained CHIPMUNK models for all test languages except for Hebrew, as Hebrew lacks gold segmentation data. Following \citet{P17-1184}, we use the default parameters for Morfessor, and 10k merge operations for BPE across languages.
We use available BPE models pre-trained on Wikipedia by \newcite{Heinzerling2018BPEmbTP}.\footnote{\url{https://github.com/bheinzerling/bpemb}} 


Two well-known word representation models, which can also be described by the general framework from Figure~\ref{fig:model_arc}, are used as insightful baselines: the subword-agnostic \textsc{sgns} model \citep{DBLP:conf/nips/MikolovSCCD13} and \texttt{fastText} (\textsc{ft})\footnote{\url{https://github.com/facebookresearch/fastText}} \citep{Q17-1010}. \textsc{ft} computes the target word embedding using addition as the composition function, while the segmentation is straightforward: the model simply generates all character n-grams of length 3 to 6 and adds them to $S_w$ along with the full word.

\paragraph{Training Setup}
Our training data for all languages is Wikipedia. We lowercase all text and replace all digits with a generic tag \textit{\#}. The statistics of the training corpora are provided in Table~\ref{tb:wikist}. 

All subword-informed variants are trained on the same data and share the same parameters for the \textsc{sgns} model.\footnote{We rely on the standard choices: $300$-dimensional subword and word embeddings, 5 training epochs, the context window size is 5, 5 negative samples, the subsampling rate of $10^{-5}$, and the minimum word frequency is 5.}
Further, we use {\sc adagrad} \citep{Duchi:2011:ASM:1953048.2021068} with a linearly decaying learning rate, and do a grid search of learning rate and batch size for each $\delta$ on the German\footnote{German has moderate morphological complexity among the five languages, so we think the hyperparameters tuned on it could be applicable to other languages.} WordSim-353 data set (\textsc{ws}; \citet{DBLP:journals/corr/LeviantR15}). The hyper-parameters are then fixed for all other languages and evaluation runs. 
Finally, we set the learning rate to 0.05 for \textit{sms} and \textit{bpe}, and 0.075 for \textit{morf}, and the batch size to 1024 for all the settings.


\begin{table}[t]
	\centering
    \def\arraystretch{0.97}
    {\footnotesize
	\begin{tabularx}{\columnwidth}{r cXX}
		Typology & Language & \#tokens & \#types \\
		\toprule
		Fusional  & English (\textsc{en}) & 600M & 900K \\		
				  & German (\textsc{de})  & 200M &  940K\\		
        \cmidrule(lr){2-4}
		Agglutinative &  Finnish (\textsc{fi})   & 66M & 600K \\
		 			  &  Turkish (\textsc{tr})   & 52M & 300K \\
        \cmidrule(lr){2-4}
		Introflexive &  Hebrew (\textsc{he})   & 90M & 410K \\
        \bottomrule
	\end{tabularx}}%
	\caption{Statistics of our Wikipedia training corpora. For faster training, we use one third of the entire Wikipedia corpus for \textsc{en} and \textsc{de}.}\label{tb:wikist} 
    \vspace{-1.5mm}
\end{table}

\subsection{Evaluation Tasks}

\paragraph{Word Similarity and Relatedness}\label{sec:ws}
These standard intrinsic evaluation tasks test the semantics of word representations \citep{pennington2014glove,Q17-1010}. The evaluations are performed using the Spearman's rank correlation score between the average of human judgement similarity scores for word pairs and the cosine similarity between two word embeddings constituting each word pair.
We use Multilingual SimLex-999 (\textsc{simlex}; \citet{Hill:2015cl,DBLP:journals/corr/LeviantR15,DBLP:journals/tacl/MrksicVSLRGKY17}) for English, German and Hebrew, each containing 999 word pairs annotated for true semantic similarity. We further evaluate embeddings on FinnSim-300 (\textsc{fs300}) produced by \citet{venekoski2017finnish} for Finnish and AnlamVer (\textsc{an}; \citet{C18-1323}) for Turkish. We also run experiments on the WordSim-353 test set (\textsc{ws}; \citet{Finkelstein:2002tois}), and its portions oriented towards true similarity (\textsc{ws-sim}) and broader relatedness (\textsc{ws-rel}) portion for English and German.
\begin{table*}[t]
	\centering
    \def\arraystretch{0.97}
    {\footnotesize
	\begin{tabularx}{\textwidth}{ll lll XX}
		\multicolumn{2}{r}{}  & Best & 2nd Best & Worst & \texttt{sgns} & \texttt{ft}\\ \cmidrule(lr){3-5} \cmidrule(lr){6-7} 
         \multirow{4}{*}{\textsc{en}} 
         &	\textsc{ws} & {\bf .656} ({\it sms.w-.st.att}) 
                        & .655 ({\it sms.w-.pp.att})
                        & .440 ({\it bpe.w-.mp.mtx})  & .634  & .643\\
         & \textsc{ws-sim}  &  {\bf .708} ({\it sms.ww.st.mtx}) 
                            & .707 ({\it sms.w-.st.att}) 
                            & .475 ({\it bpe.w-.mp.att})  & .702  & .706\\ 
         &	\textsc{ws-rel} &  {\bf .625} ({\it sms.w-.p-.add}) 
                            & .620 ({\it sms.w-.st.att})
                            & .438 ({\it bpe.w-.mp.mtx}) & .579 & .586\\
         & \textsc{simlex}  & .283 ({\it sms.ww.p-.add}) 
                            & .282 ({\it morf.w-.p-.add})
                            & .182 ({\it bpe.w-.mp.add})  & .300  & {\bf .307}\\ \cmidrule(lr){3-5} \cmidrule(lr){6-7}
                            
        \multirow{4}{*}{\textsc{de}} 
        &  \textsc{ws}  & {\bf .633} ({\it sms.ww.pp.add}) 
                        & {\bf .633} ({\it sms.ww.p-.add})
                        & .328 ({\it bpe.w-.mp.mtx}) & .596  & .624\\ 
        & \textsc{ws-sim}   & .673 ({\it sms.ww.pp.add}) 
                            & .668 ({\it sms.ww.p-.add})
                            & .363 ({\it bpe.w-.mp.add}) & .669  & {\bf .677}\\
        & \textsc{ws-rel}   & {\bf .616} ({\it sms.ww.p-.add}) 
                            & .610 ({\it sms.ww.pp.add})
                            & .332 ({\it bpe.w-.mp.mtx}) & .530  & .590\\
        & \textsc{simlex}   & {\bf .401} ({\it sms.ww.p-.add})
                            & .398 ({\it sms.ww.pp.add}) 
                            & .189 ({\it bpe.w-.mp.mtx}) & .359 & .393 \\ \cmidrule(lr){3-5} \cmidrule(lr){6-7}
                            
        \textsc{fi} 
        & \textsc{fs300}    & .259 ({\it sms.w-.p-.add}) 
                            & .258 ({\it sms.w-.pp.add})
                            & .123 ({\it morf.ww.mp.mtx}) & .211 & {\bf .279}\\ \cmidrule(lr){3-5} \cmidrule(lr){6-7}

        \multirow{2}{*}{\textsc{tr}} 
        & \textsc{an-sim}   & {\bf .355} ({\it bpe.ww.mp.add}) 
                            & .325 ({\it bpe.ww.mp.att})
                            & .112 ({\it bpe.ww.pp.mtx}) & .232 & .271\\
        & \textsc{an-rel}   & .459 ({\it bpe.ww.mp.add}) 
                            & .444 ({\it sms.w-.pp.add})
                            & .273 ({\it morf.ww.mp.att}) & .183 & {\bf .520}\\ \cmidrule(lr){3-5} \cmidrule(lr){6-7}
                            
         \textsc{he}    
         & \textsc{simlex}  & .338 ({\it bpe.ww.pp.add})
                            & .338 ({\it bpe.ww.pp.mtx})
                            & .128 ({\it bpe.w-.mp.mtx}) & .379 & {\bf .388}\\ \cmidrule(lr){3-5} \cmidrule(lr){6-6} \cmidrule(lr){7-7} \cmidrule(lr){3-5} \cmidrule(lr){6-7}
         
        \textsc{en} 
        & \textsc{card} & {\bf .370} ({\it sms.ww.pp.add}) 
                        & .328 ({\it sms.w-.pp.mtx}) 
                        & .000 ({\it bpe.w-.pp.add}) & .009 & .249\\
		\bottomrule       
	\end{tabularx}}%
    \vspace{-1mm}
     \caption{Results on word similarity and relatedness across languages. The highest score for each row is in bold, and we choose randomly in case of a tie. All scores are obtained after computing the embeddings of OOV words.} \label{tb:ws} 
     \vspace{-2mm}
\end{table*}
\begin{table}[t]
	\centering
    \def\arraystretch{0.97}
    {\footnotesize
	\begin{tabularx}{\columnwidth}{ll cccc}
		\multicolumn{2}{r}{}  & \multicolumn{2}{c}{Dev set} &
        \multicolumn{2}{c}{Test set}\\ \cmidrule(lr){3-4} \cmidrule(lr){5-6}
		\multicolumn{2}{r}{}  & UAS & LAS & UAS & LAS\\ \cmidrule(lr){3-4} \cmidrule(lr){5-6}
        
        \multirow{5}{*}{\textsc{en}} 
         & {\it sms.w-.mp.mtx} & {\bf 92.3} & 90.3 & 92.0 & 90.1\\
         & {\it bpe.ww.p-.mtx} & {\bf 92.3} & {\bf 90.4} & {\bf 92.1} & 90.0\\
         & \texttt{sgns} & {\bf 92.3} & {\bf 90.4}	& 91.9	& 89.8\\
         & \texttt{ft} & {\bf 92.3} & 90.3 & {\bf 92.1} & {\bf 90.2}\\ \cmidrule(lr){3-6}
         
        \multirow{5}{*}{\textsc{de}} 
         & {\it bpe.ww.pp.add} & 91.2 & 87.7 & {\bf 89.6} & {\bf 84.7}\\
         & {\it bpe.ww.mp.mtx} & 91.2 & 87.7 & 89.4 & {\bf 84.7}\\
         & \texttt{sgns}	& 91.4 & {\bf 87.9} & 89.3 & 84.4\\
         & \texttt{ft}		& {\bf 91.6} & {\bf 87.9} & 89.1 & 84.4\\ \cmidrule(lr){3-6}
         
        \multirow{5}{*}{\textsc{fi}}     
         & {\it bpe.ww.mp.add} & {\bf 89.9} & {\bf 86.9} & {\bf 90.7} & {\bf 87.4}\\
         & {\it sms.w-.pp.add} & 89.3 & 86.1 & 90.5 & 87.1\\
         & \texttt{sgns}	& {\bf 88.9} & 85.6 & 89.5 & 86.2\\
         & \texttt{ft}	& 89.7 & {\bf 86.9} & 90.4 & 87.1\\ \cmidrule(lr){3-6}
         
        \multirow{5}{*}{\textsc{tr}} 
         & {\it sms.ww.mp.add} & 71.1 & {\bf 63.5} & 72.8 & 64.7\\
         & {\it sms.w-.mp.att} & 70.5 & 62.5 & 72.7 & 64.5\\
         & \texttt{sgns}	& 70.5	& 62.5	& 72.2	& 63.5\\
         & \texttt{ft}	& {\bf 71.2} & 63.3 & {\bf 73.1} & {\bf 65.1}\\ \cmidrule(lr){3-6}
         
        \multirow{5}{*}{\textsc{he}} 
         & {\it morf.w-.pp.mtx} & 92.3 & 89.5 & 91.3 & 88.5\\
         & {\it morf.ww.p-.add} & 92.3 & 89.5 & 91.2 & 88.5\\
         & \texttt{sgns}	& 92.4	& {\bf 89.8}	& {\bf 91.5}	& {\bf 88.7}\\
         & \texttt{ft}	& {\bf 92.6} & 89.7 & 91.2 & 88.3\\
		\bottomrule       
	\end{tabularx}}%
    \vspace{-1mm}
     \caption{Results on the dependency parsing task. The two best configurations are selected according to LAS.}\label{tb:ps} 
     \vspace{-1mm}
\end{table}

Finally, to analyze the importance of subword information for learning embeddings of rare words, we evaluate on the recently released CARD-660 dataset (\textsc{card}; \citet{DBLP:conf/emnlp/PilehvarKPC18}) for English, annotated for true semantic similarity.

\paragraph{Dependency Parsing}
Next, we use the syntactic dependency parsing task to analyze the importance of subword information for syntactically-driven downstream applications. For all test languages, we rely on the standard Universal Dependencies treebanks (UD v2.2; \citet{DBLP:conf/lrec/NivreMGGHMMPPST16}). We use subword-informed word embeddings from different configurations to initialize the deep biaffine parser of \citet{DBLP:journals/corr/DozatM16} which has shown competitive performance in shared tasks \citep{K17-3002} and among other parsing models \citep{DBLP:conf/ijcnlp/MaH17,D17-1002,ma18acl}.\footnote{\url{https://github.com/tdozat/Parser-v2}}
We use default settings for the biaffine parser for all experimental runs

\paragraph{Fine-Grained Entity Typing}
The task is to map entities, which could comprise more than one entity token, to predefined entity types \citep{yaghoobzadeh-schutze:2015:EMNLP}. 
It is a suitable semi-semantic task to test our subword models, as the subwords of entities usually carry some semantic information from which the entity types can be inferred.
For example, {\it Lincolnshire} will belong to \texttt{/location/county} as {\it -shire} is a suffix that strongly indicates a location.
We rely on an entity typing dataset of \citet{Heinzerling2018BPEmbTP} built for over 250 languages by obtaining entity mentions from Wikidata \citep{42240} and their associated FIGER-based entity types \cite{Ling:2012:FER:2900728.2900742}: there only exists a one-to-one mapping between the entity and one of the 112 FIGER types.

We randomly sample the data to obtain a train/dev/test split with the size of 60k/20k/20k for all languages. For evaluation we extend the RNN-based model of \citet{Heinzerling2018BPEmbTP}, where they stacked all the subwords of entity tokens into a flattened sequence: we use the hierarchical embedding composition instead. For each entity token, we first compute its word embeddings with our subword configurations,\footnote{Although it is true that case information can be very important to the task, we conform to \citet{Heinzerling2018BPEmbTP} lowercasing all letters.} then feed the word embeddings of entity tokens to a bidirectional LSTM with 2 hidden layers of size 512, followed by a projection layer which predicts the entity type.



\section{Results and Analysis}
\begin{table*}[t]
	\centering
    \def\arraystretch{0.91}
    {\footnotesize
	\begin{tabularx}{\textwidth}{l lll XX}
		\multicolumn{1}{r}{}  & Best & 2nd Best & Worst & \texttt{sgns} & \texttt{ft} \\ \cmidrule(lr){2-4} \cmidrule(lr){5-6} 
         \textsc{en}   &  {\bf 55.70} ({\it bpe.ww.mp.add}) 
         			   &  55.68 ({\it morf.w-.pp.att}) 
                       &  51.15 ({\it bpe.w-.p-.att})  
                       & 51.00  & 55.15 \\ 
              
         \textsc{de}   &  54.06 ({\it morf.w-.pp.add}) 
         			   &  54.01 ({\it morf.w-.pp.att}) 
                       &  50.21 ({\it bpe.w-.p-.att})  
                       & 50.14  & {\bf 54.55} 
                       \\ \cmidrule(lr){2-4} \cmidrule(lr){5-6} 
                      
         \textsc{fi}   &  \textbf{57.41} ({\it morf.w-.pp.add}) 
         			   &  57.38 ({\it morf.w-.pp.mtx}) 
                       &  52.18 ({\it bpe.w-.p-.att})  
                       & 49.87  & 57.18 \\ 
                       
         \textsc{tr}   &  {\bf 56.31} ({\it morf.w-.pp.add}) 
         			   &  56.14 ({\it morf.w-.pp.mtx}) 
                       &  46.97 ({\it bpe.w-.pp.mtx})  
                       & 54.35  & 54.62 \\ \cmidrule(lr){2-4} \cmidrule(lr){5-6} 
        
         \textsc{he}   &  {\bf 60.34}  ({\it morf.w-.pp.att}) 
         			   &  60.09 ({\it morf.ww.pp.add}) 
                       &  51.31 ({\it bpe.w-.p-.att})  
                       &  54.55 & 59.09 \\ 
		\bottomrule       
	\end{tabularx}}%
    \vspace{-2mm}
     \caption{Test accuracy, the evaluation metric used by \citet{Heinzerling2018BPEmbTP}, on the fine-grained entity typing task. The results are averaged over 5 runs with random seeds.} \label{tb:et} 
\end{table*}
\begin{figure*}[!t]
	\centering
    \includegraphics[width=0.99\textwidth, trim={2cm 3cm 2cm 4cm}, clip]{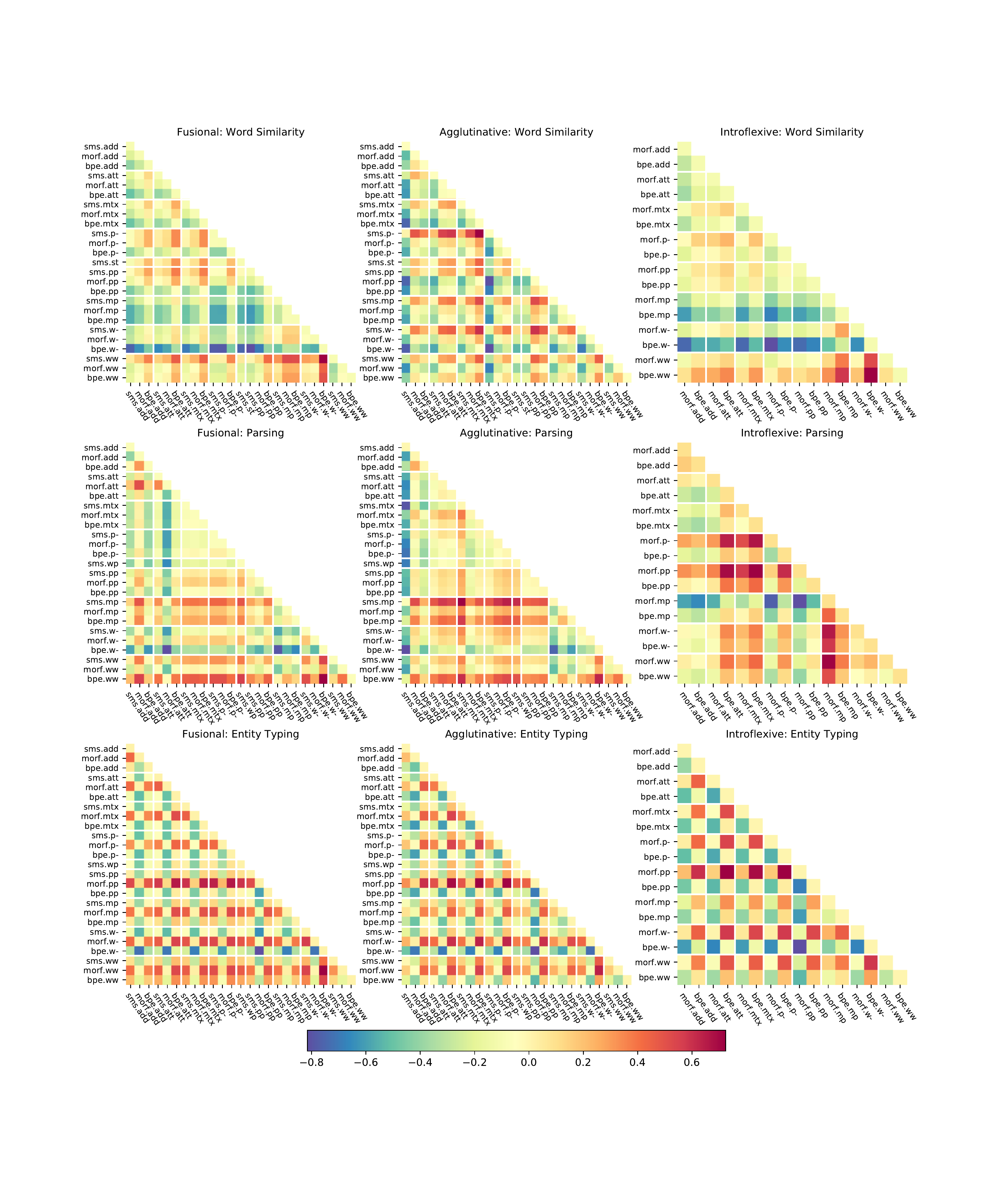}
    \vspace{-2mm}
    \caption{Comparisons of different configurations across tasks grouped by language types. 
    The value in each pixel block is the percentage rank of the row configuration minus that of column configuration.
    For example, in word similarity for fusional language, the rank for the row {\it sms.ww} is $0.787$ and $0.071$ for column {\it bpe.w-}, and the block value is $0.716$.
    The higher the value, the better the performance of the row compared to the column configuration.} \label{fig:config_comp}
    \vspace{-1mm}
\end{figure*}

To get a better grasp of the overall performance without overloading the tables, we focus on reporting two best configurations and the worst configuration for each task and language from the total of 60 configurations, except for Hebrew with 36 configurations, where there is no gold segmentation data for training \textit{sms} model. We also analyze the effects of different configurations on different tasks based on language typology. The entire analysis revolves around the key questions Q1-Q3 posed in the introduction. The reader is encouraged to refer to the supplementary material for the complete results. 

Tables~\ref{tb:ws},~\ref{tb:ps}, and~\ref{tb:et} summarize the main results on word similarity and relatedness, dependency parsing and entity typing, respectively. In addition, the comparisons of different configurations across tasks and language types are shown in Figure~\ref{fig:config_comp} (as well as Figure~\ref{fig:config_comp2} to~\ref{fig:worm_wordemb} in the supplementary material). There, we center the comparison around two crucial components: segmentation and composition. The value in each pixel block is the percentage rank of the row configuration minus that of column configuration.
We compute such percentage ranks by performing three levels of averaging over: 1) all related datasets for the same task; 2) all sub-configurations that entail the configuration in question; 3) all languages from the same language types.


\paragraph{Q1. Tasks and Languages} Regarding the absolute performance of our subword-informed configurations, we notice that they outperform \textsc{sgns} and \textsc{ft} in 3/5 languages on average, and for 8/13 datasets on word similarity and relatedness. The gains are more prominent over the subword-agnostic \textsc{sgns} model and for morphologically richer languages such as Finnish and Turkish. The results on the two other tasks are also very competitive, with strong performance reported especially on the entity typing task. This clearly indicates the importance of integrating subword-level information into word vectors. A finer-grained comparative analysis shows that best-performing configurations vary greatly across different languages. 

The comparative analysis across tasks also suggests that there is no single configuration that outperforms the others in all three tasks, although certain patterns in the results emerge. For instance, the supervised segmentation (\textit{sms}) is very useful for word similarity and relatedness (seen in Figure~\ref{fig:worm_seg}). This result is quite intuitive: {\it sms} is trained according to the readily available gold standard morphological segmentations. However, \textit{sms} is less useful for entity typing, where almost all best-performing configurations are based on \textit{morf} (see also Figure~\ref{fig:worm_seg}). This result is also interpretable: {\it morf} is a conservative segmenter that captures longer subwords, and is not distracted by short and nonsensical subwords (like \textit{bpe}) that have no contribution to the prediction.\footnote{For example, {\it sms} and {\it bpe} both split ``Valberg" (\texttt{/location/city}) and ``Robert Valberg" (\texttt{/person/actor}) with a suffix ``berg". Since ``berg" could represent both a place or a person, it is not useful alone as a suffix to predict the correct entity type, whereas {\it morf} does not split the word and makes the prediction solely based on surrounding entity tokens.} The worst configurations on word similarity are based on \textit{bpe}: its aggressive segmentation often results in non-interpretable or nonsensical subwords which are unable to recover useful semantic information. Due to the same reason, the results in all three tasks indicate that the best configurations with {\it bpe} are always coupled with {\it ww}, and the worst are obtained with {\it w-} (i.e., without the inclusion of the full word).

The results on parsing reveal similar performances for a spectrum of heterogeneous configurations. In other words, while the chosen configuration is still important, its impact on performance of state-of-the-art dependency parsers \cite{Kipperwasser:2016tacl,DBLP:journals/corr/DozatM16} is decreased, as such parsers are heavily parametrized multi-component methods (e.g., besides word embeddings they rely on biLSTMs, intra-sentence attention, character representations).\footnote{We also experimented with removing token features such as POS tags and character embeddings in some settings, but we observed similar trends in the final results.} Therefore, a larger space of subword-informed configurations for word representations leads to optimal or near-optimal results. However, \textit{sms} seems to yield highest scores on average in agglutinative languages (see also Figure~\ref{fig:config_comp}). 

Figure~\ref{fig:config_comp} clearly demonstrates that, apart from entity typing heatmaps (the third row) which show very similar trends over different language types, the patterns for different tasks and language types tend to vary in general. Similarly, other figures in the supplementary material also show diverging trends across languages representing different language types.

\paragraph{Q2 and Q3. Configurations} 
As mentioned, different tasks reach high scores with different segmentation and composition components. The crucial components for word similarity are {\it sms} and {\it ww}, and {\it sms} is generally better than {\it morf} and {\it bpe} in fusional and agglutinative languages (see Figure~\ref{fig:worm_seg} and~\ref{fig:worm_wordemb}). The presence of {\it ww} is desired in this task as also found by \newcite{Q17-1010}: it enhances the information provided by the segmentation. As discussed before, the best configurations with {\it bpe} are always coupled with {\it ww}, and the worst with {\it w-}. {\it ww} is less important for the more conservative {\it morf}  where the information stored in {\it ww} can be fully recovered from the generated subwords. Interestingly, {\it pp} and {\it mp} do not have positive effects on this task for fusional and introflexive languages, but they seem to resonate well with agglutinative languages, and they are useful for the two other tasks (seen in Figure~\ref{fig:worm_posemb}). In general, position embeddings have shown potential benefits in all tasks, where they selectively emphasize or filter subwords according to their positions. {\it pp} is extremely useful in entity typing for all languages, because it indicates the root position.

Concerning composition functions, {\it add} still remains an extremely robust choice across tasks and languages. Surprisingly, the more sophisticated self-attention composition prevails only on a handful of datasets: compare the results with \textit{add} vs. \textit{att} and \textit{mtx}. In fact, the worst configurations mostly use \textit{att} and \textit{mtx} (see also Figure~\ref{fig:worm_compf}). In sum, our results suggest that, when unsure, \textit{add} is by far the most robust choice for the subword composition function. Further, morphotactic tags encoded in subword embeddings ({\it st}) seem to be only effective combined with self-attention in word similarity and relatedness.  These findings call for further investigation in future work, along with the inclusion of finer-grained morphotactic tags into the proposed modeling framework.

\paragraph{Further Discussion}
A recurring theme of this study is that subword-informed configurations are largely task- and language-dependent. We can extract multiple examples from the reported results affirming this conjecture. For instance, in fusional and agglutinative languages {\it mp} is critical to the model on dependency parsing, while for Hebrew, an introflexive language, {\it mp} is among the most detrimental components on the same task. Further, for Turkish word similarity {\it bpe.ww} outperforms {\it sms.ww}: due to affix concatenation in Turkish, {\it sms} produces many affixes with only syntactic functions that bring noise to the task. Interestingly, \textsc{sgns} performs well in Hebrew on parsing and word similarity: it shows that it is still difficult for linear segmentation methods to capture non-concatenative morphology.

Finally, fine-tuning subword-informed representations seems especially beneficial for rare word semantics: our best configuration outperforms \textsc{ft} by 0.111 on \textsc{card}, and even surpasses all the state-of-the-art models on the rare word similarity task, as reported by \citet{DBLP:conf/emnlp/PilehvarKPC18}. We hope that our findings on the \textsc{card} dataset will motivate further work on building more accurate representations for rare and unseen words \cite{Bhatia:2016emnlp,Herbelot:2017emnlp,Schick:2018arxiv} by learning more effective and more informed components of subword-informed configurations.

\section{Conclusion}
We have presented a general framework for learning subword-informed word representations which has been used to perform a systematic analysis of 60 different subword-aware configurations for 5 typologically diverse languages across 3 diverse tasks. The large space of presented results has allowed us to analyze the main properties of subword-informed representation learning: we have demonstrated that different components of the framework such as segmentation and composition methods, or the use of position embeddings, have to be carefully tuned to yield improved performance across different tasks and languages. We hope that this study will guide the development of new subword-informed word representation architectures. Code is available at: \url{https://github.com/cambridgeltl/sw_study}.


\section*{Acknowledgments}
This work is supported by the ERC Consolidator Grant LEXICAL (no 648909). We thank the reviewers for their insightful comments, and Roi Reichart for many fruitful discussions.

\bibliographystyle{acl_natbib}
\bibliography{mybib}

\appendix
\section{Supplemental Material}
In this supplementary material we present the full results of different configurations of our subword-informed representations and baseline models in three tasks across five languages.
For details and notations of different model configurations, please refer to the original paper.

Since \texttt{fastText} (\textsc{ft}) package cannot generate each subword embedding for a given word, we also implemented our own version (\texttt{our\_ft}) within the same subword-informed training framework.
We trained \texttt{our\_ft} on all the five languages with the same training corpus, and kept the same hyperparameters as \textsc{ft} package except that we used 1024 as batch size for faster training.
We show that our implementation yields comparable results with the original \textsc{ft} in word similarity and relatedness and parsing across languages.
For entity typing, we experimented with the extended hierarchical composition architecture described in the paper on \texttt{our\_ft}, i.e., using subword embeddings to compute word representations and updating subword embeddings of \texttt{our\_ft} during training, and only use word embeddings as input for \textsc{sgns} and \textsc{ft} and directly update them.
Additionally, we also experimented with some of our model configurations with character n-grams as our segmentation methods, but did not observe performance gains. 
Due to its large parameter space (usually 2-4 times larger than \textsc{sgns}) and the computational limits, we leave it as our future work.

\begin{figure*}[b]
	\centering
    \includegraphics[width=1.0\textwidth, trim={2cm 0 2cm 0.5cm}, clip]{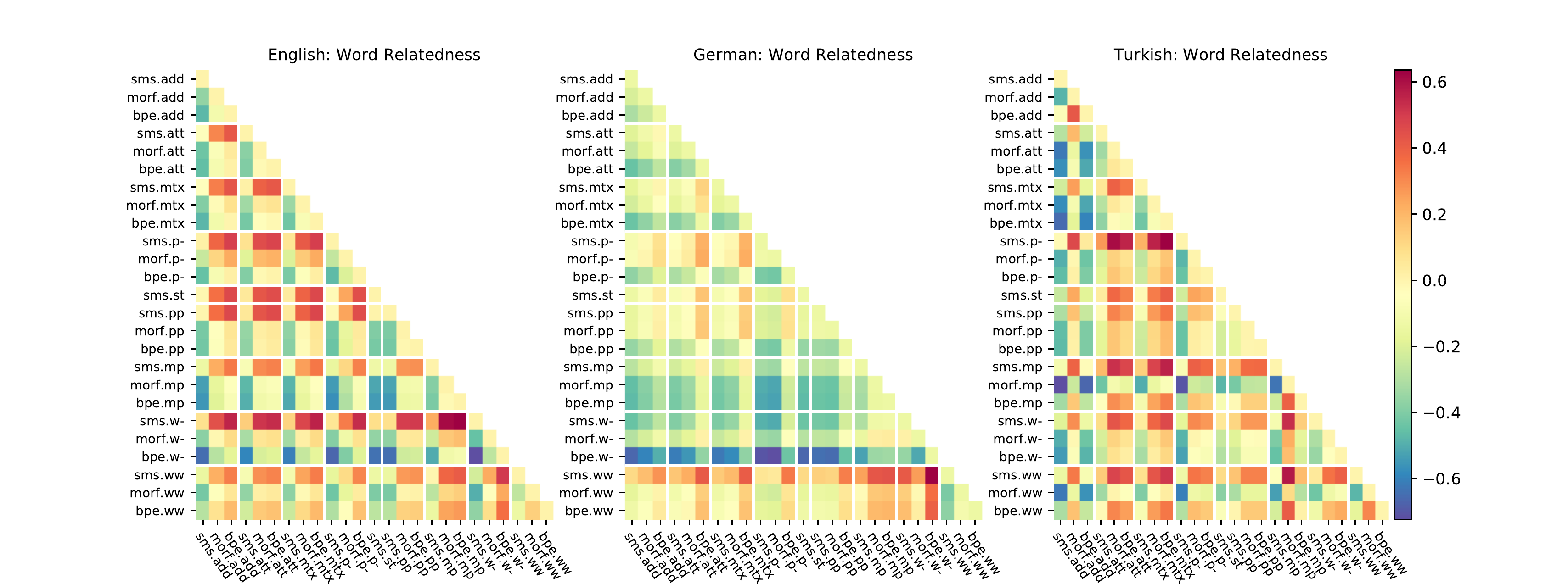}
    \caption{Comparisons of word relatedness tasks for English, German and Turkish.} \label{fig:config_comp2}
\end{figure*}

\begin{figure*}[b]
	\centering
    \includegraphics[width=1.0\textwidth, trim={2cm 0 2cm 1cm}, clip]{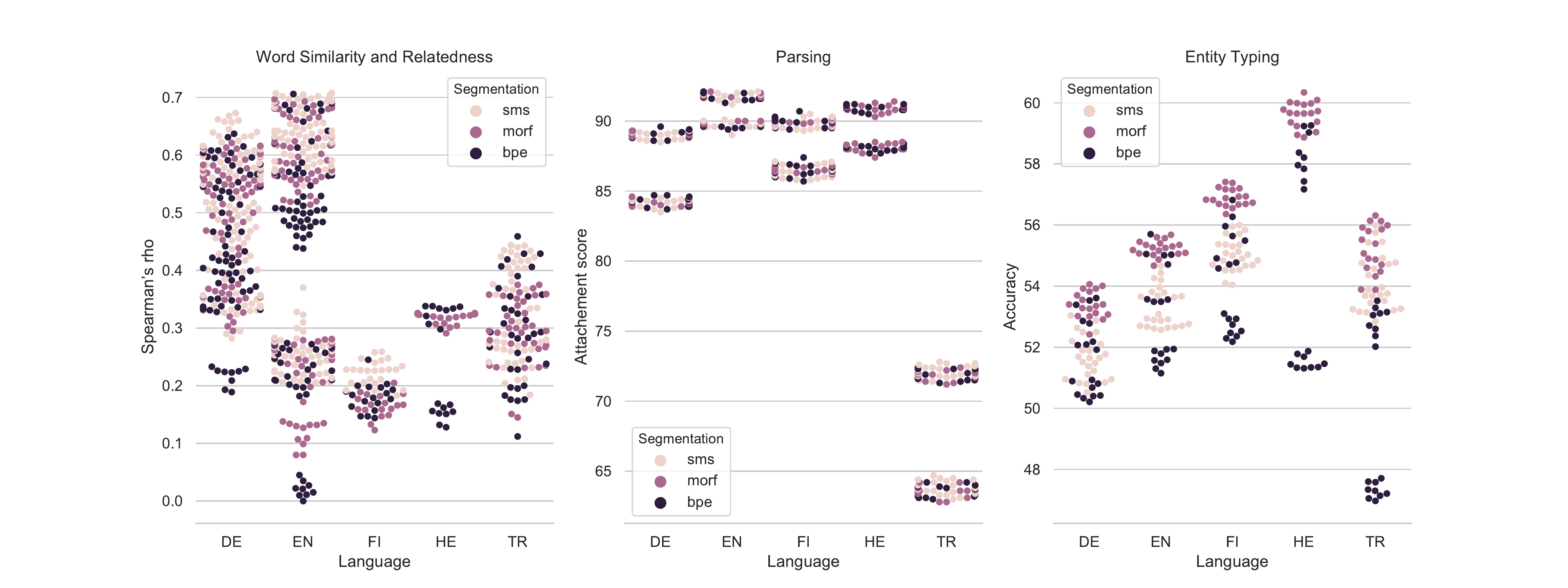}
    \caption{Results with different segmentation methods.} \label{fig:worm_seg}
\end{figure*}

\begin{figure*}[b]
	\centering
    \includegraphics[width=1.0\textwidth, trim={2cm 0 2cm 1cm}, clip]{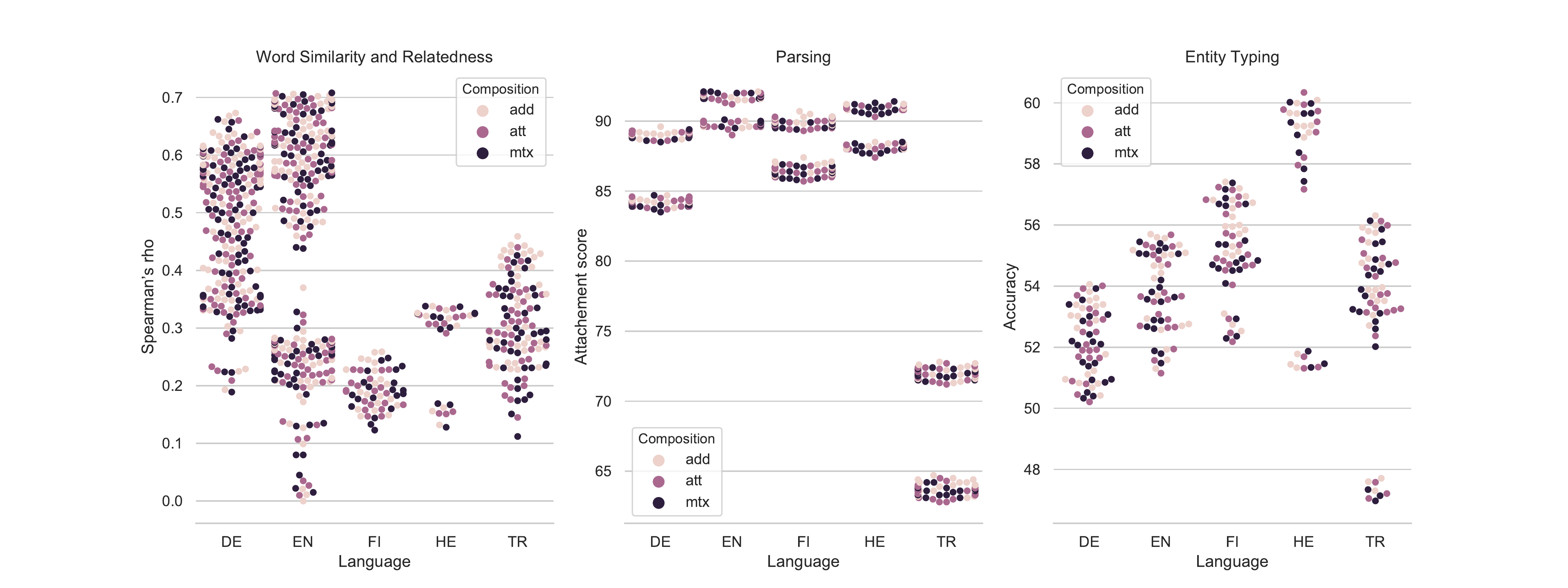}
    \caption{Results with different composition functions.} \label{fig:worm_compf}
\end{figure*}

\begin{figure*}[b]
	\centering
    \includegraphics[width=1.0\textwidth, trim={2cm 0 2cm 1cm}, clip]{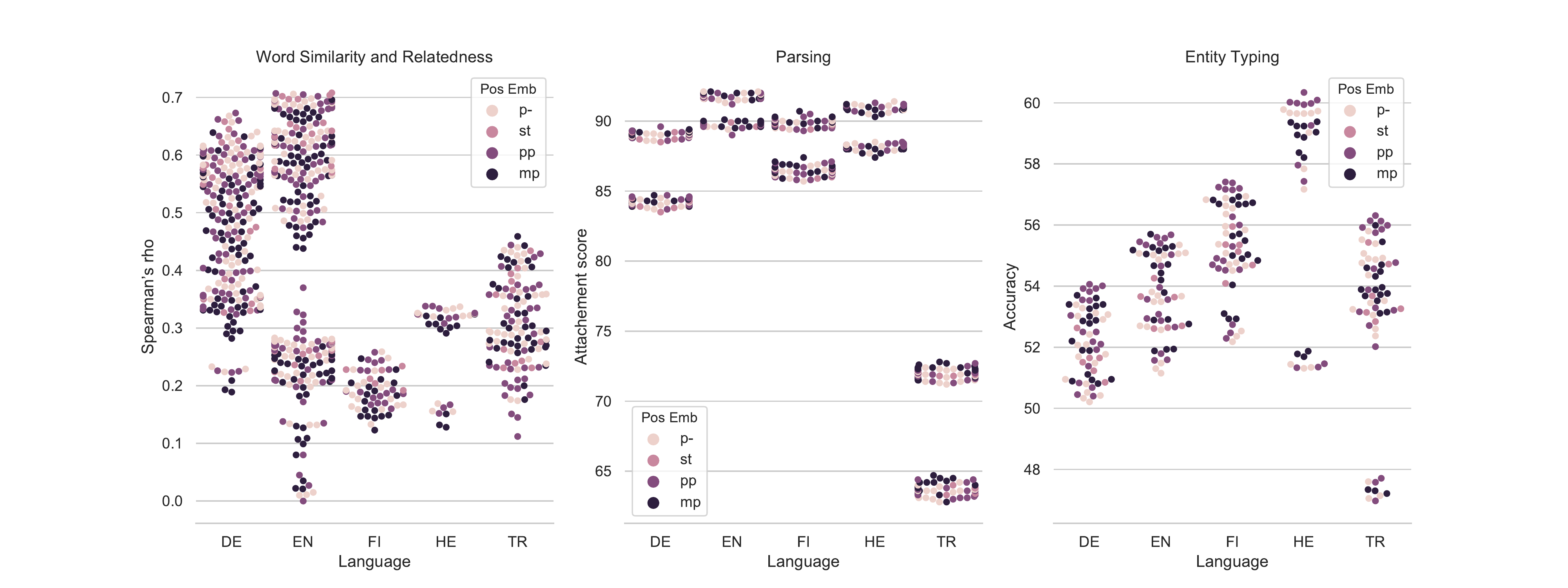}
    \caption{Results with different position embeddings.} \label{fig:worm_posemb}
\end{figure*}

\begin{figure*}[b]
	\centering
    \includegraphics[width=1.0\textwidth, trim={2cm 0 2cm 1cm}, clip]{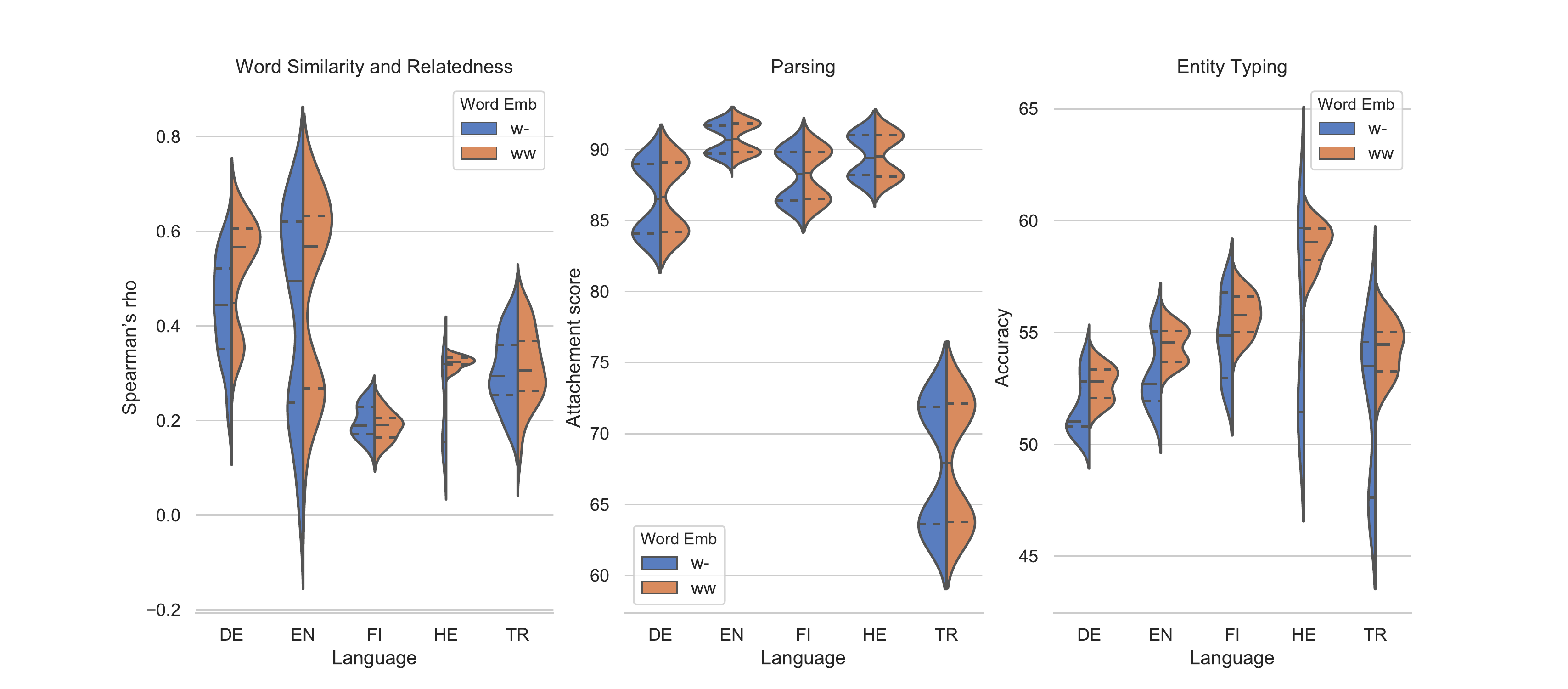}
    \caption{Results w/o word token embedding.} \label{fig:worm_wordemb}
\end{figure*}

Table \ref{tb:ws_sms}, \ref{tb:ws_morf} and \ref{tb:ws_bpe} show the results of word similarity and relatedness based on different segmentation methods, and Table \ref{tb:ps_sms}, \ref{tb:ps_morf} and \ref{tb:ps_bpe} for the results of dependency parsing.
Table \ref{tb:et2} shows the accuracy on development and test set for entity typing across all five languages.
Figure~\ref{fig:config_comp2} shows the configuration comparisons in languages with available datasets for word relatedness. 
Figure~\ref{fig:worm_seg} to~\ref{fig:worm_wordemb} show the results on all downstream tasks across languages when each test data point is categorized according to some specific component.
Specifically, Figure~\ref{fig:worm_seg} shows the results when only different segmentation methods are considered, and  Figure~\ref{fig:worm_compf},~\ref{fig:worm_posemb} and~\ref{fig:worm_wordemb} show the results on varying composition functions, position embedding types and whether including the word token embedding, respectively.


\begin{table*}[t]
	\centering
    \def\arraystretch{0.85}
    {\footnotesize
	\begin{tabularx}{\textwidth}{l XXXXXXXX}
	    \toprule
	    & \textsc{simlex} & \textsc{ws} & \textsc{ws-rel} & \textsc{ws-sim} & \textsc{fs300} & \textsc{an-rel} & \textsc{an-sim} & \textsc{card}\\ \cmidrule(lr){2-9} 
		
    \multirow{2}{*}{{\it sms.w-.p-.add}}
    & 0.267(\textsc{en}) & 0.650 & 0.625 & 0.688 & 0.259(\textsc{fi}) & 0.277(\textsc{tr}) & 0.432(\textsc{tr})& 0.206(\textsc{en})\\
    & 0.350(\textsc{de}) & 0.517 & 0.490 & 0.586\\\cmidrule(lr){2-9}
    \multirow{2}{*}{{\it sms.w-.st.add}}
    & 0.267 & 0.654 & 0.619 & 0.704 & 0.228 & 0.276 & 0.434 & 0.208\\
    & 0.350 & 0.509 & 0.475 & 0.577\\\cmidrule(lr){2-9}
    \multirow{2}{*}{{\it sms.w-.pp.add}}
    & 0.270 & 0.645 & 0.615 & 0.693 & 0.258 & 0.261 & 0.444 & 0.294\\
    & 0.332 & 0.494 & 0.467 & 0.563\\\cmidrule(lr){2-9}
    \multirow{2}{*}{{\it sms.w-.mp.add}}
    & 0.207 & 0.635 & 0.598 & 0.683 & 0.247 & 0.269 & 0.424 & 0.213\\
    & 0.295 & 0.437 & 0.401 & 0.507\\\cmidrule(lr){2-9}
    \multirow{2}{*}{{\it sms.ww.p-.add}}
    & 0.283 & 0.640 & 0.589 & 0.696 & 0.226 & 0.273 & 0.435 & 0.267\\
    & 0.401 & 0.633 & 0.616 & 0.668\\\cmidrule(lr){2-9}
    \multirow{2}{*}{{\it sms.ww.st.add}}
    & 0.277 & 0.637 & 0.576 & 0.697 & 0.204 & 0.239 & 0.404 & 0.269\\
    & 0.385 & 0.622 & 0.601 & 0.658\\\cmidrule(lr){2-9}
    \multirow{2}{*}{{\it sms.ww.pp.add}}
    & 0.282 & 0.632 & 0.572 & 0.702 & 0.240 & 0.240 & 0.423 & 0.370\\
    & 0.398 & 0.633 & 0.610 & 0.673\\\cmidrule(lr){2-9}
    \multirow{2}{*}{{\it sms.ww.mp.add}}
    & 0.240 & 0.633 & 0.579 & 0.696 & 0.211 & 0.279 & 0.443 & 0.268\\
    & 0.353 & 0.599 & 0.557 & 0.639\\\cmidrule(lr){2-9}
    \multirow{2}{*}{{\it sms.w-.p-.att}}
    & 0.250 & 0.642 & 0.613 & 0.684 & 0.228 & 0.289 & 0.416 & 0.244\\
    & 0.339 & 0.518 & 0.478 & 0.598\\\cmidrule(lr){2-9}
    \multirow{2}{*}{{\it sms.w-.st.att}}
    & 0.255 & 0.656 & 0.620 & 0.697 & 0.227 & 0.263 & 0.358 & 0.236\\
    & 0.327 & 0.512 & 0.468 & 0.583\\\cmidrule(lr){2-9}
    \multirow{2}{*}{{\it sms.w-.pp.att}}
    & 0.252 & 0.655 & 0.619 & 0.707 & 0.227 & 0.212 & 0.334 & 0.323\\
    & 0.324 & 0.500 & 0.456 & 0.570\\\cmidrule(lr){2-9}
    \multirow{2}{*}{{\it sms.w-.mp.att}}
    & 0.206 & 0.629 & 0.593 & 0.674 & 0.205 & 0.263 & 0.405 & 0.219\\
    & 0.290 & 0.452 & 0.425 & 0.517\\\cmidrule(lr){2-9}
    \multirow{2}{*}{{\it sms.ww.p-.att}}
    & 0.277 & 0.642 & 0.591 & 0.698 & 0.192 & 0.311 & 0.440 & 0.251\\
    & 0.360 & 0.605 & 0.571 & 0.640\\\cmidrule(lr){2-9}
    \multirow{2}{*}{{\it sms.ww.st.att}}
    & 0.261 & 0.632 & 0.570 & 0.693 & 0.212 & 0.243 & 0.366 & 0.260\\
    & 0.342 & 0.588 & 0.549 & 0.625\\\cmidrule(lr){2-9}
    \multirow{2}{*}{{\it sms.ww.pp.att}}
    & 0.271 & 0.630 & 0.584 & 0.695 & 0.228 & 0.204 & 0.367 & 0.310\\
    & 0.357 & 0.608 & 0.567 & 0.660\\\cmidrule(lr){2-9}
    \multirow{2}{*}{{\it sms.ww.mp.att}}
    & 0.220 & 0.616 & 0.547 & 0.682 & 0.191 & 0.241 & 0.376 & 0.222\\
    & 0.340 & 0.569 & 0.526 & 0.613\\\cmidrule(lr){2-9}
    \multirow{2}{*}{{\it sms.w-.p-.mtx}}
    & 0.245 & 0.639 & 0.602 & 0.694 & 0.248 & 0.260 & 0.404 & 0.220\\
    & 0.336 & 0.505 & 0.457 & 0.585\\\cmidrule(lr){2-9}
    \multirow{2}{*}{{\it sms.w-.st.mtx}}
    & 0.252 & 0.648 & 0.617 & 0.702 & 0.234 & 0.231 & 0.381 & 0.224\\
    & 0.330 & 0.506 & 0.456 & 0.578\\\cmidrule(lr){2-9}
    \multirow{2}{*}{{\it sms.w-.pp.mtx}}
    & 0.260 & 0.638 & 0.606 & 0.694 & 0.244 & 0.184 & 0.335 & 0.328\\
    & 0.331 & 0.488 & 0.429 & 0.564\\\cmidrule(lr){2-9}
    \multirow{2}{*}{{\it sms.w-.mp.mtx}}
    & 0.211 & 0.620 & 0.599 & 0.671 & 0.228 & 0.265 & 0.417 & 0.226\\
    & 0.282 & 0.447 & 0.417 & 0.506\\\cmidrule(lr){2-9}
    \multirow{2}{*}{{\it sms.ww.p-.mtx}}
    & 0.279 & 0.631 & 0.575 & 0.696 & 0.206 & 0.280 & 0.426 & 0.251\\
    & 0.357 & 0.615 & 0.578 & 0.657\\\cmidrule(lr){2-9}
    \multirow{2}{*}{{\it sms.ww.st.mtx}}
    & 0.277 & 0.642 & 0.581 & 0.708 & 0.192 & 0.231 & 0.394 & 0.261\\
    & 0.355 & 0.603 & 0.561 & 0.645\\\cmidrule(lr){2-9}
    \multirow{2}{*}{{\it sms.ww.pp.mtx}}
    & 0.273 & 0.640 & 0.588 & 0.705 & 0.226 & 0.210 & 0.368 & 0.300\\
    & 0.357 & 0.611 & 0.575 & 0.662\\\cmidrule(lr){2-9}
    \multirow{2}{*}{{\it sms.ww.mp.mtx}}
    & 0.253 & 0.622 & 0.565 & 0.684 & 0.183 & 0.275 & 0.414 & 0.260\\
    & 0.339 & 0.598 & 0.578 & 0.632\\\cmidrule(lr){2-9}
        \multirow{3}{*}{\texttt{sgns}}
    & 0.300 & 0.634 & 0.579 & 0.702 & 0.211 & 0.232 & 0.183 & 0.009\\
    & 0.359 & 0.596 & 0.530 & 0.669\\\cmidrule(lr){2-9}
    \multirow{3}{*}{\texttt{ft}}
    & 0.307 & 0.643 & 0.586 & 0.706 & 0.279 & 0.271 & 0.520 & 0.249\\
    & 0.393 & 0.624 & 0.590 & 0.677\\\cmidrule(lr){2-9}
    \multirow{3}{*}{\texttt{our\_ft}}
    & 0.265 & 0.629 & 0.574 & 0.702 & 0.241 & 0.295 & 0.523 & 0.281\\
    & 0.380 & 0.610 & 0.596 & 0.649\\
	\bottomrule       
    \end{tabularx}}
     \caption{Results on word similarity and relatedness across languages for CHIPMUNK ({\it sms}). All scores are obtained after computing the embeddings of OOV words.} \label{tb:ws_sms} 
\end{table*}

\begin{table*}[t]
	\centering
	    \def\arraystretch{0.85}
    {\footnotesize
	\begin{tabularx}{\textwidth}{l XXXXXXXX}
	    \toprule
	    & \textsc{simlex} & \textsc{ws} & \textsc{ws-rel} & \textsc{ws-sim} & \textsc{fs300} & \textsc{an-rel} & \textsc{an-sim} & \textsc{card}\\ \cmidrule(lr){2-9} 
    \multirow{3}{*}{{\it morf.w-.p-.add}}
    & 0.282(\textsc{en}) & 0.629 & 0.574 & 0.691 & 0.166(\textsc{fi}) & 0.273(\textsc{tr}) & 0.359(\textsc{tr}) & 0.134(\textsc{en})\\
    & 0.368(\textsc{de}) & 0.577 & 0.557 & 0.616\\
    & 0.322(\textsc{he})\\\cmidrule(lr){2-9}
    \multirow{3}{*}{{\it morf.w-.pp.add}}
    & 0.279 & 0.617 & 0.562 & 0.680 & 0.189 & 0.232 & 0.375 & 0.172\\
    & 0.349 & 0.549 & 0.530 & 0.594\\
    & 0.317\\\cmidrule(lr){2-9}
    \multirow{3}{*}{{\it morf.w-.mp.add}}
    & 0.245 & 0.597 & 0.529 & 0.664 & 0.185 & 0.302 & 0.351 & 0.127\\
    & 0.303 & 0.484 & 0.439 & 0.550\\
    & 0.324\\\cmidrule(lr){2-9}
    \multirow{3}{*}{{\it morf.ww.p-.add}}
    & 0.273 & 0.632 & 0.573 & 0.697 & 0.159 & 0.264 & 0.339 & 0.198\\
    & 0.376 & 0.583 & 0.569 & 0.614\\
    & 0.326\\\cmidrule(lr){2-9}
    \multirow{3}{*}{{\it morf.ww.pp.add}}
    & 0.277 & 0.616 & 0.557 & 0.684 & 0.182 & 0.232 & 0.375 & 0.262\\
    & 0.361 & 0.573 & 0.549 & 0.609\\
    & 0.324\\\cmidrule(lr){2-9}
    \multirow{3}{*}{{\it morf.ww.mp.add}}
    & 0.244 & 0.589 & 0.520 & 0.666 & 0.149 & 0.267 & 0.300 & 0.099\\
    & 0.331 & 0.515 & 0.484 & 0.574\\
    & 0.304\\\cmidrule(lr){2-9}
    \multirow{3}{*}{{\it morf.w-.p-.att}}
    & 0.281 & 0.623 & 0.565 & 0.682 & 0.159 & 0.295 & 0.362 & 0.132\\
    & 0.354 & 0.560 & 0.536 & 0.603\\
    & 0.323\\\cmidrule(lr){2-9}
    \multirow{3}{*}{{\it morf.w-.pp.att}}
    & 0.276 & 0.619 & 0.555 & 0.682 & 0.170 & 0.235 & 0.358 & 0.138\\
    & 0.352 & 0.556 & 0.537 & 0.597\\
    & 0.308\\\cmidrule(lr){2-9}
    \multirow{3}{*}{{\it morf.w-.mp.att}}
    & 0.246 & 0.586 & 0.519 & 0.661 & 0.158 & 0.274 & 0.304 & 0.107\\
    & 0.310 & 0.495 & 0.453 & 0.548\\
    & 0.291\\\cmidrule(lr){2-9}
    \multirow{3}{*}{{\it morf.ww.p-.att}}
    & 0.271 & 0.622 & 0.568 & 0.686 & 0.167 & 0.292 & 0.346 & 0.218\\
    & 0.372 & 0.574 & 0.545 & 0.621\\
    & 0.321\\\cmidrule(lr){2-9}
    \multirow{3}{*}{{\it morf.ww.pp.att}}
    & 0.274 & 0.613 & 0.549 & 0.680 & 0.167 & 0.145 & 0.312 & 0.206\\
    & 0.364 & 0.565 & 0.537 & 0.601\\
    & 0.319\\\cmidrule(lr){2-9}
    \multirow{3}{*}{{\it morf.ww.mp.att}}
    & 0.234 & 0.586 & 0.514 & 0.666 & 0.147 & 0.270 & 0.273 & 0.109\\
    & 0.327 & 0.499 & 0.469 & 0.562\\
    & 0.307\\\cmidrule(lr){2-9}
    \multirow{3}{*}{{\it morf.w-.p-.mtx}}
    & 0.280 & 0.632 & 0.573 & 0.693 & 0.133 & 0.300 & 0.357 & 0.132\\
    & 0.348 & 0.564 & 0.544 & 0.602\\
    & 0.320\\\cmidrule(lr){2-9}
    \multirow{3}{*}{{\it morf.w-.pp.mtx}}
    & 0.275 & 0.621 & 0.562 & 0.693 & 0.186 & 0.231 & 0.369 & 0.135\\
    & 0.355 & 0.549 & 0.530 & 0.582\\
    & 0.325\\\cmidrule(lr){2-9}
    \multirow{3}{*}{{\it morf.w-.mp.mtx}}
    & 0.254 & 0.602 & 0.536 & 0.671 & 0.205 & 0.235 & 0.295 & 0.130\\
    & 0.295 & 0.488 & 0.427 & 0.542\\
    & 0.301\\\cmidrule(lr){2-9}
    \multirow{3}{*}{{\it morf.ww.p-.mtx}}
    & 0.275 & 0.624 & 0.558 & 0.688 & 0.177 & 0.292 & 0.356 & 0.222\\
    & 0.371 & 0.577 & 0.545 & 0.611\\
    & 0.325\\\cmidrule(lr){2-9}
    \multirow{3}{*}{{\it morf.ww.pp.mtx}}
    & 0.272 & 0.617 & 0.558 & 0.684 & 0.160 & 0.151 & 0.321 & 0.080\\
    & 0.359 & 0.592 & 0.575 & 0.613\\
    & 0.326\\\cmidrule(lr){2-9}
    \multirow{3}{*}{{\it morf.ww.mp.mtx}}
    & 0.238 & 0.591 & 0.522 & 0.674 & 0.123 & 0.279 & 0.326 & 0.080\\
    & 0.320 & 0.500 & 0.465 & 0.559\\
    & 0.318\\\cmidrule(lr){2-9}
    \multirow{3}{*}{\texttt{sgns}}
    & 0.300 & 0.634 & 0.579 & 0.702 & 0.211 & 0.232 & 0.183 & 0.009\\
    & 0.359 & 0.596 & 0.530 & 0.669\\
    & 0.379\\\cmidrule(lr){2-9}
    \multirow{3}{*}{\texttt{ft}}
    & 0.307 & 0.643 & 0.586 & 0.706 & 0.279 & 0.271 & 0.520 & 0.249\\
    & 0.393 & 0.624 & 0.590 & 0.677\\
    & 0.388\\\cmidrule(lr){2-9}
    \multirow{3}{*}{\texttt{our\_ft}}
    & 0.265 & 0.629 & 0.574 & 0.702 & 0.241 & 0.295 & 0.523 & 0.281\\
    & 0.380 & 0.610 & 0.596 & 0.649\\
    & 0.354\\
		\bottomrule       
    \end{tabularx}}
     \caption{Results on word similarity and relatedness across languages for Morfessor ({\it morf}). All scores are obtained after computing the embeddings of OOV words.} \label{tb:ws_morf} 
\end{table*}

\begin{table*}[t]
	\centering
	\def\arraystretch{0.85}
    {\footnotesize
	\begin{tabularx}{\textwidth}{l XXXXXXXX}
	    \toprule
	    & \textsc{simlex} & \textsc{ws} & \textsc{ws-rel} & \textsc{ws-sim} & \textsc{fs300} & \textsc{an-rel} & \textsc{an-sim} & \textsc{card}\\ \cmidrule(lr){2-9} 
    \multirow{3}{*}{{\it bpe.w-.p-.add}}
    & 0.209(\textsc{en}) & 0.488 & 0.499 & 0.508 & 0.177(\textsc{fi}) & 0.228(\textsc{tr}) & 0.390(\textsc{tr}) & 0.011(\textsc{en})\\
    & 0.229(\textsc{de}) & 0.416 & 0.442 & 0.474\\
    & 0.156(\textsc{he})\\\cmidrule(lr){2-9}
    \multirow{3}{*}{{\it bpe.w-.pp.add}}
    & 0.209 & 0.474 & 0.484 & 0.490 & 0.170 & 0.229 & 0.407 & 0.000\\
    & 0.225 & 0.404 & 0.417 & 0.467\\
    & 0.162\\\cmidrule(lr){2-9}
    \multirow{3}{*}{{\it bpe.w-.mp.add}}
    & 0.182 & 0.460 & 0.478 & 0.484 & 0.157 & 0.283 & 0.406 & 0.021\\
    & 0.193 & 0.331 & 0.351 & 0.363\\
    & 0.132\\\cmidrule(lr){2-9}
    \multirow{3}{*}{{\it bpe.ww.p-.add}}
    & 0.274 & 0.642 & 0.577 & 0.706 & 0.203 & 0.293 & 0.429 & 0.262\\
    & 0.378 & 0.597 & 0.582 & 0.631\\
    & 0.331\\\cmidrule(lr){2-9}
    \multirow{3}{*}{{\it bpe.ww.pp.add}}
    & 0.265 & 0.631 & 0.571 & 0.689 & 0.245 & 0.267 & 0.429 & 0.273\\
    & 0.365 & 0.615 & 0.595 & 0.637\\
    & 0.338\\\cmidrule(lr){2-9}
    \multirow{3}{*}{{\it bpe.ww.mp.add}}
    & 0.236 & 0.587 & 0.519 & 0.658 & 0.147 & 0.355 & 0.459 & 0.240\\
    & 0.348 & 0.566 & 0.538 & 0.608\\
    & 0.321\\\cmidrule(lr){2-9}
    \multirow{3}{*}{{\it bpe.w-.p-.att}}
    & 0.216 & 0.506 & 0.504 & 0.529 & 0.200 & 0.176 & 0.288 & 0.010\\
    & 0.233 & 0.386 & 0.409 & 0.422\\
    & 0.155\\\cmidrule(lr){2-9}
    \multirow{3}{*}{{\it bpe.w-.pp.att}}
    & 0.210 & 0.503 & 0.507 & 0.528 & 0.190 & 0.183 & 0.331 & 0.027\\
    & 0.226 & 0.383 & 0.396 & 0.428\\
    & 0.152\\\cmidrule(lr){2-9}
    \multirow{3}{*}{{\it bpe.w-.mp.att}}
    & 0.197 & 0.456 & 0.462 & 0.475 & 0.173 & 0.200 & 0.308 & 0.035\\
    & 0.209 & 0.398 & 0.422 & 0.467\\
    & 0.151\\\cmidrule(lr){2-9}
    \multirow{3}{*}{{\it bpe.ww.p-.att}}
    & 0.265 & 0.621 & 0.566 & 0.687 & 0.197 & 0.246 & 0.358 & 0.257\\
    & 0.350 & 0.555 & 0.525 & 0.591\\
    & 0.334\\\cmidrule(lr){2-9}
    \multirow{3}{*}{{\it bpe.ww.pp.att}}
    & 0.270 & 0.618 & 0.570 & 0.678 & 0.187 & 0.197 & 0.314 & 0.228\\
    & 0.333 & 0.567 & 0.536 & 0.604\\
    & 0.334\\\cmidrule(lr){2-9}
    \multirow{3}{*}{{\it bpe.ww.mp.att}}
    & 0.255 & 0.613 & 0.564 & 0.681 & 0.147 & 0.325 & 0.419 & 0.226\\
    & 0.338 & 0.545 & 0.526 & 0.584\\
    & 0.307\\\cmidrule(lr){2-9}
    \multirow{3}{*}{{\it bpe.w-.p-.mtx}}
    & 0.202 & 0.486 & 0.485 & 0.512 & 0.184 & 0.174 & 0.293 & 0.015\\
    & 0.224 & 0.394 & 0.414 & 0.433\\
    & 0.169\\\cmidrule(lr){2-9}
    \multirow{3}{*}{{\it bpe.w-.pp.mtx}}
    & 0.198 & 0.504 & 0.500 & 0.547 & 0.179 & 0.195 & 0.355 & 0.045\\
    & 0.223 & 0.399 & 0.396 & 0.462\\
    & 0.167\\\cmidrule(lr){2-9}
    \multirow{3}{*}{{\it bpe.w-.mp.mtx}}
    & 0.185 & 0.440 & 0.438 & 0.476 & 0.144 & 0.176 & 0.282 & 0.022\\
    & 0.189 & 0.328 & 0.332 & 0.372\\
    & 0.128\\\cmidrule(lr){2-9}
    \multirow{3}{*}{{\it bpe.ww.p-.mtx}}
    & 0.267 & 0.624 & 0.565 & 0.690 & 0.164 & 0.238 & 0.334 & 0.272\\
    & 0.354 & 0.573 & 0.543 & 0.604\\
    & 0.337\\\cmidrule(lr){2-9}
    \multirow{3}{*}{{\it bpe.ww.pp.mtx}}
    & 0.263 & 0.621 & 0.569 & 0.677 & 0.193 & 0.112 & 0.290 & 0.210\\
    & 0.337 & 0.553 & 0.500 & 0.608\\
    & 0.338\\\cmidrule(lr){2-9}
    \multirow{3}{*}{{\it bpe.ww.mp.mtx}}
    & 0.260 & 0.620 & 0.564 & 0.681 & 0.198 & 0.257 & 0.369 & 0.247\\
    & 0.336 & 0.546 & 0.514 & 0.596\\
    & 0.298\\\cmidrule(lr){2-9}
    \multirow{3}{*}{\texttt{sgns}}
    & 0.300 & 0.634 & 0.579 & 0.702 & 0.211 & 0.232 & 0.183 & 0.009\\
    & 0.359 & 0.596 & 0.530 & 0.669\\
    & 0.379\\\cmidrule(lr){2-9}
    \multirow{3}{*}{\texttt{ft}}
    & 0.307 & 0.643 & 0.586 & 0.706 & 0.279 & 0.271 & 0.520 & 0.249\\
    & 0.393 & 0.624 & 0.590 & 0.677\\
    & 0.388\\\cmidrule(lr){2-9}
    \multirow{3}{*}{\texttt{our\_ft}}
    & 0.265 & 0.629 & 0.574 & 0.702 & 0.241 & 0.295 & 0.523 & 0.281\\
    & 0.380 & 0.610 & 0.596 & 0.649\\
    & 0.354\\
		\bottomrule       
    \end{tabularx}}
     \caption{Results on word similarity and relatedness across languages for BPE ({\it bpe}). All scores are obtained after computing the embeddings of OOV words.} \label{tb:ws_bpe}
\end{table*}

\begin{table*}[t]
	\centering
	\def\arraystretch{0.85}
    {\footnotesize
	\begin{tabularx}{\textwidth}{l XXXXXXXX}
	    \toprule
	    & \textsc{dev(en)} & \textsc{test} 
	    & \textsc{dev(de)} & \textsc{test}
	    & \textsc{dev(fi)} & \textsc{test}
	    & \textsc{dev(tr)} & \textsc{test}\\ \cmidrule(lr){2-3} \cmidrule(lr){4-5}\cmidrule(lr){6-7}\cmidrule(lr){8-9}
    \multirow{2}{*}{{\it sms.w-.p-.add}}
    & 92.2(\textsc{uas}) & 91.5 & 91.3 & 89.0 & 88.8 & 89.6 & 70.0 & 71.9\\
    & 90.3(\textsc{las}) & 89.6 & 87.7 & 84.2 & 85.5 & 86.1 & 62.1 & 63.7\\\cmidrule(lr){2-3}\cmidrule(lr){4-5}\cmidrule(lr){6-7}\cmidrule(lr){8-9}
    \multirow{2}{*}{{\it sms.w-.st.add}}
    & 92.4 & 91.7 & 91.1 & 88.7 & 89.1 & 89.7 & 71.0 & 72.0\\
    & 90.5 & 89.7 & 87.6 & 83.8 & 86.0 & 86.6 & 62.9 & 63.8\\\cmidrule(lr){2-3}\cmidrule(lr){4-5}\cmidrule(lr){6-7}\cmidrule(lr){8-9}
    \multirow{2}{*}{{\it sms.w-.pp.add}}
    & 92.2 & 91.9 & 91.4 & 88.7 & 89.3 & 90.5 & 70.9 & 72.7\\
    & 90.2 & 89.9 & 87.8 & 84.0 & 86.1 & 87.1 & 62.8 & 64.4\\\cmidrule(lr){2-3}\cmidrule(lr){4-5}\cmidrule(lr){6-7}\cmidrule(lr){8-9}
    \multirow{2}{*}{{\it sms.w-.mp.add}}
    & 92.2 & 91.9 & 91.1 & 89.1 & 89.0 & 89.9 & 70.5 & 72.6\\
    & 90.3 & 90.0 & 87.6 & 84.3 & 85.8 & 86.6 & 62.6 & 64.5\\\cmidrule(lr){2-3}\cmidrule(lr){4-5}\cmidrule(lr){6-7}\cmidrule(lr){8-9}
    \multirow{2}{*}{{\it sms.ww.p-.add}}
    & 92.2 & 92.1 & 91.6 & 89.1 & 89.4 & 90.1 & 71.0 & 71.8\\
    & 90.2 & 90.0 & 88.1 & 84.3 & 86.4 & 86.9 & 62.8 & 63.6\\\cmidrule(lr){2-3}\cmidrule(lr){4-5}\cmidrule(lr){6-7}\cmidrule(lr){8-9}
    \multirow{2}{*}{{\it sms.ww.st.add}}
    & 92.4 & 91.9 & 91.3 & 89.2 & 89.4 & 89.9 & 71.0 & 72.0\\
    & 90.5 & 89.8 & 87.9 & 84.5 & 86.2 & 86.5 & 63.0 & 64.0\\\cmidrule(lr){2-3}\cmidrule(lr){4-5}\cmidrule(lr){6-7}\cmidrule(lr){8-9}
    \multirow{2}{*}{{\it sms.ww.pp.add}}
    & 92.5 & 91.9 & 91.3 & 88.9 & 89.5 & 90.3 & 70.7 & 72.5\\
    & 90.5 & 90.0 & 87.8 & 84.2 & 86.5 & 87.0 & 63.0 & 64.4\\\cmidrule(lr){2-3}\cmidrule(lr){4-5}\cmidrule(lr){6-7}\cmidrule(lr){8-9}
    \multirow{2}{*}{{\it sms.ww.mp.add}}
    & 92.2 & 91.7 & 91.4 & 89.2 & 89.4 & 90.3 & 71.1 & 72.8\\
    & 90.3 & 89.6 & 87.8 & 84.4 & 86.3 & 87.1 & 63.5 & 64.7\\\cmidrule(lr){2-3}\cmidrule(lr){4-5}\cmidrule(lr){6-7}\cmidrule(lr){8-9}
    \multirow{2}{*}{{\it sms.w-.p-.att}}
    & 92.1 & 91.7 & 91.3 & 88.9 & 89.0 & 89.6 & 71.3 & 72.4\\
    & 90.2 & 89.6 & 87.7 & 83.9 & 85.9 & 86.4 & 63.1 & 63.8\\\cmidrule(lr){2-3}\cmidrule(lr){4-5}\cmidrule(lr){6-7}\cmidrule(lr){8-9}
    \multirow{2}{*}{{\it sms.w-.st.att}}
    & 92.2 & 91.6 & 91.4 & 89.0 & 88.8 & 89.4 & 70.0 & 71.9\\
    & 90.3 & 89.6 & 87.7 & 84.2 & 85.6 & 86.0 & 62.2 & 63.4\\\cmidrule(lr){2-3}\cmidrule(lr){4-5}\cmidrule(lr){6-7}\cmidrule(lr){8-9}
    \multirow{2}{*}{{\it sms.w-.pp.att}}
    & 91.4 & 91.2 & 91.1 & 89.0 & 89.2 & 89.7 & 70.7 & 71.4\\
    & 89.4 & 89.0 & 87.6 & 84.3 & 85.7 & 86.1 & 62.6 & 63.0\\\cmidrule(lr){2-3}\cmidrule(lr){4-5}\cmidrule(lr){6-7}\cmidrule(lr){8-9}
    \multirow{2}{*}{{\it sms.w-.mp.att}}
    & 92.1 & 91.8 & 91.3 & 89.0 & 89.0 & 90.0 & 70.5 & 72.7\\
    & 90.2 & 89.8 & 87.7 & 84.1 & 85.9 & 86.6 & 62.5 & 64.5\\\cmidrule(lr){2-3}\cmidrule(lr){4-5}\cmidrule(lr){6-7}\cmidrule(lr){8-9}
    \multirow{2}{*}{{\it sms.ww.p-.att}}
    & 92.1 & 91.5 & 91.2 & 89.3 & 89.2 & 89.8 & 71.2 & 72.3\\
    & 90.2 & 89.6 & 87.7 & 84.5 & 85.9 & 86.4 & 63.2 & 64.0\\\cmidrule(lr){2-3}\cmidrule(lr){4-5}\cmidrule(lr){6-7}\cmidrule(lr){8-9}
    \multirow{2}{*}{{\it sms.ww.st.att}}
    & 92.3 & 91.9 & 91.3 & 89.2 & 88.7 & 89.7 & 70.4 & 72.2\\
    & 90.4 & 89.9 & 87.8 & 84.3 & 85.5 & 86.2 & 62.3 & 63.8\\\cmidrule(lr){2-3}\cmidrule(lr){4-5}\cmidrule(lr){6-7}\cmidrule(lr){8-9}
    \multirow{2}{*}{{\it sms.ww.pp.att}}
    & 92.2 & 91.8 & 91.5 & 89.0 & 88.5 & 89.3 & 71.0 & 72.2\\
    & 90.3 & 89.8 & 87.9 & 84.4 & 84.9 & 85.8 & 62.7 & 63.7\\\cmidrule(lr){2-3}\cmidrule(lr){4-5}\cmidrule(lr){6-7}\cmidrule(lr){8-9}
    \multirow{2}{*}{{\it sms.ww.mp.att}}
    & 92.3 & 91.9 & 91.5 & 89.1 & 88.6 & 89.5 & 70.7 & 72.5\\
    & 90.5 & 90.0 & 88.0 & 84.3 & 85.4 & 86.0 & 62.9 & 64.3\\\cmidrule(lr){2-3}\cmidrule(lr){4-5}\cmidrule(lr){6-7}\cmidrule(lr){8-9}
    \multirow{2}{*}{{\it sms.w-.p-.mtx}}
    & 92.0 & 91.9 & 91.5 & 88.6 & 89.0 & 89.7 & 70.5 & 71.7\\
    & 90.1 & 89.8 & 87.9 & 83.7 & 85.9 & 86.2 & 62.5 & 63.5\\\cmidrule(lr){2-3}\cmidrule(lr){4-5}\cmidrule(lr){6-7}\cmidrule(lr){8-9}
    \multirow{2}{*}{{\it sms.w-.st.mtx}}
    & 92.1 & 91.8 & 91.1 & 88.5 & 89.0 & 89.5 & 69.8 & 71.8\\
    & 90.2 & 89.7 & 87.8 & 83.5 & 85.6 & 85.9 & 61.8 & 63.3\\\cmidrule(lr){2-3}\cmidrule(lr){4-5}\cmidrule(lr){6-7}\cmidrule(lr){8-9}
    \multirow{2}{*}{{\it sms.w-.pp.mtx}}
    & 92.3 & 91.7 & 91.2 & 89.1 & 88.9 & 89.6 & 70.7 & 72.0\\
    & 90.4 & 89.7 & 87.6 & 84.2 & 85.9 & 86.3 & 62.6 & 63.1\\\cmidrule(lr){2-3}\cmidrule(lr){4-5}\cmidrule(lr){6-7}\cmidrule(lr){8-9}
    \multirow{2}{*}{{\it sms.w-.mp.mtx}}
    & 92.3 & 92.0 & 91.4 & 89.1 & 89.4 & 90.2 & 70.7 & 72.4\\
    & 90.3 & 90.1 & 87.8 & 84.3 & 86.0 & 86.9 & 62.5 & 64.2\\\cmidrule(lr){2-3}\cmidrule(lr){4-5}\cmidrule(lr){6-7}\cmidrule(lr){8-9}
    \multirow{2}{*}{{\it sms.ww.p-.mtx}}
    & 92.2 & 91.6 & 91.4 & 89.1 & 89.2 & 89.8 & 70.5 & 71.9\\
    & 90.4 & 89.6 & 87.9 & 84.3 & 86.0 & 86.4 & 62.5 & 63.6\\\cmidrule(lr){2-3}\cmidrule(lr){4-5}\cmidrule(lr){6-7}\cmidrule(lr){8-9}
    \multirow{2}{*}{{\it sms.ww.st.mtx}}
    & 92.2 & 91.6 & 91.4 & 88.7 & 89.1 & 89.8 & 71.0 & 71.8\\
    & 90.4 & 89.7 & 87.8 & 83.9 & 85.8 & 86.5 & 62.7 & 63.3\\\cmidrule(lr){2-3}\cmidrule(lr){4-5}\cmidrule(lr){6-7}\cmidrule(lr){8-9}
    \multirow{2}{*}{{\it sms.ww.pp.mtx}}
    & 92.3 & 92.0 & 91.2 & 89.0 & 88.7 & 89.4 & 70.4 & 71.8\\
    & 90.4 & 89.9 & 87.7 & 84.2 & 85.2 & 85.8 & 62.7 & 63.6\\\cmidrule(lr){2-3}\cmidrule(lr){4-5}\cmidrule(lr){6-7}\cmidrule(lr){8-9}
    \multirow{2}{*}{{\it sms.ww.mp.mtx}}
    & 92.3 & 92.0 & 91.4 & 88.9 & 88.8 & 89.6 & 71.0 & 71.5\\
    & 90.4 & 89.9 & 87.8 & 84.0 & 85.2 & 85.9 & 62.7 & 63.3\\\cmidrule(lr){2-3}\cmidrule(lr){4-5}\cmidrule(lr){6-7}\cmidrule(lr){8-9}
    
        \multirow{2}{*}{\texttt{sgns}}
        & 92.3 & 91.9 & 91.4 & 89.3 & 88.9 & 89.5 & 70.5 & 72.2\\
        & 90.4 & 89.8 & 87.9 & 84.4 & 85.6 & 86.2 & 62.5 & 63.5\\\cmidrule(lr){2-3}\cmidrule(lr){4-5}\cmidrule(lr){6-7}\cmidrule(lr){8-9}
        \multirow{2}{*}{\texttt{ft}}
        & 92.3 & 92.1 & 91.6 & 89.1 & 89.7 & 90.4 & 71.2 & 73.1\\
        & 90.3 & 90.2 & 87.9 & 84.4 & 86.9 & 87.1 & 63.3 & 65.1\\\cmidrule(lr){2-3}\cmidrule(lr){4-5}\cmidrule(lr){6-7}\cmidrule(lr){8-9}
        \multirow{2}{*}{\texttt{our\_ft}}
        & 92.7 & 92.2 & 91.6 & 89.6 & 90.5 & 90.9 & 71.5 & 73.0\\
        & 90.7 & 90.2 & 87.9 & 84.9 & 87.6 & 87.8 & 63.5 & 64.9\\
		\bottomrule       
    \end{tabularx}}    
     \caption{Results on dependency parsing across languages for CHIPMUNK ({\it sms}).} \label{tb:ps_sms} 
\end{table*}

\begin{table*}[t]
	\centering
	\def\arraystretch{0.85}
    {\footnotesize
	\begin{tabularx}{\textwidth}{l XXXXXXXXXX}
	    \toprule
	    & \textsc{dev(en)} & \textsc{test} 
	    & \textsc{dev(de)} & \textsc{test}
	    & \textsc{dev(fi)} & \textsc{test}
	    & \textsc{dev(tr)} & \textsc{test}
	    & \textsc{dev(he)} & \textsc{test}\\ \cmidrule(lr){2-3} \cmidrule(lr){4-5}\cmidrule(lr){6-7}\cmidrule(lr){8-9}\cmidrule(lr){10-11}
    \multirow{2}{*}{{\it morf.w-.p-.add}}
    & 92.3 & 91.7 & 91.1 & 89.0 & 89.0 & 90.0 & 70.7 & 71.8 & 92.5 & 91.0\\
    & 90.4 & 89.8 & 87.7 & 84.2 & 85.8 & 86.6 & 62.4 & 63.3 & 89.6 & 88.3\\\cmidrule(lr){2-3}\cmidrule(lr){4-5}\cmidrule(lr){6-7}\cmidrule(lr){8-9}\cmidrule(lr){10-11}
    \multirow{2}{*}{{\it morf.w-.pp.add}}
    & 92.5 & 91.8 & 91.4 & 89.0 & 89.2 & 89.8 & 70.3 & 71.3 & 92.3 & 91.2\\
    & 90.6 & 89.8 & 87.6 & 84.1 & 86.0 & 86.4 & 62.2 & 63.1 & 89.6 & 88.4\\\cmidrule(lr){2-3}\cmidrule(lr){4-5}\cmidrule(lr){6-7}\cmidrule(lr){8-9}\cmidrule(lr){10-11}
    \multirow{2}{*}{{\it morf.w-.mp.add}}
    & 92.2 & 91.9 & 91.4 & 89.0 & 89.2 & 89.8 & 71.2 & 71.7 & 92.1 & 90.8\\
    & 90.2 & 90.0 & 87.8 & 84.1 & 85.9 & 86.4 & 63.2 & 63.6 & 89.2 & 88.0\\\cmidrule(lr){2-3}\cmidrule(lr){4-5}\cmidrule(lr){6-7}\cmidrule(lr){8-9}\cmidrule(lr){10-11}
    \multirow{2}{*}{{\it morf.ww.p-.add}}
    & 92.2 & 91.8 & 91.4 & 89.0 & 89.4 & 89.6 & 70.8 & 72.2 & 92.3 & 91.2\\
    & 90.3 & 89.7 & 87.8 & 84.1 & 86.2 & 86.2 & 62.9 & 64.2 & 89.5 & 88.5\\\cmidrule(lr){2-3}\cmidrule(lr){4-5}\cmidrule(lr){6-7}\cmidrule(lr){8-9}\cmidrule(lr){10-11}
    \multirow{2}{*}{{\it morf.ww.pp.add}}
    & 92.3 & 91.7 & 91.3 & 88.9 & 88.8 & 89.9 & 70.6 & 71.9 & 92.2 & 91.0\\
    & 90.4 & 89.8 & 87.6 & 84.1 & 85.7 & 86.5 & 62.5 & 63.4 & 89.6 & 88.1\\\cmidrule(lr){2-3}\cmidrule(lr){4-5}\cmidrule(lr){6-7}\cmidrule(lr){8-9}\cmidrule(lr){10-11}
    \multirow{2}{*}{{\it morf.ww.mp.add}}
    & 92.3 & 91.7 & 91.2 & 89.0 & 89.3 & 89.8 & 70.5 & 72.4 & 92.1 & 90.9\\
    & 90.4 & 89.6 & 87.6 & 84.2 & 86.0 & 86.4 & 62.8 & 64.2 & 89.2 & 88.0\\\cmidrule(lr){2-3}\cmidrule(lr){4-5}\cmidrule(lr){6-7}\cmidrule(lr){8-9}\cmidrule(lr){10-11}
    \multirow{2}{*}{{\it morf.w-.p-.att}}
    & 92.2 & 91.8 & 91.2 & 89.1 & 89.1 & 90.2 & 70.3 & 72.1 & 92.3 & 91.1\\
    & 90.3 & 89.7 & 87.6 & 83.9 & 85.9 & 86.6 & 61.9 & 63.6 & 89.6 & 88.3\\\cmidrule(lr){2-3}\cmidrule(lr){4-5}\cmidrule(lr){6-7}\cmidrule(lr){8-9}\cmidrule(lr){10-11}
    \multirow{2}{*}{{\it morf.w-.pp.att}}
    & 92.3 & 92.0 & 91.5 & 89.3 & 89.1 & 89.7 & 71.0 & 72.2 & 92.6 & 91.2\\
    & 90.3 & 89.9 & 87.9 & 84.6 & 85.9 & 86.4 & 63.1 & 63.6 & 89.7 & 88.4\\\cmidrule(lr){2-3}\cmidrule(lr){4-5}\cmidrule(lr){6-7}\cmidrule(lr){8-9}\cmidrule(lr){10-11}
    \multirow{2}{*}{{\it morf.w-.mp.att}}
    & 92.0 & 91.8 & 91.4 & 89.1 & 89.1 & 89.9 & 70.2 & 71.5 & 92.1 & 90.3\\
    & 90.0 & 89.8 & 87.7 & 84.3 & 85.7 & 86.4 & 62.1 & 62.8 & 89.2 & 87.4\\\cmidrule(lr){2-3}\cmidrule(lr){4-5}\cmidrule(lr){6-7}\cmidrule(lr){8-9}\cmidrule(lr){10-11}
    \multirow{2}{*}{{\it morf.ww.p-.att}}
    & 92.2 & 91.9 & 91.4 & 89.1 & 89.4 & 89.8 & 70.0 & 71.2 & 92.2 & 91.1\\
    & 90.2 & 89.8 & 88.0 & 84.2 & 86.0 & 86.5 & 62.2 & 62.8 & 89.3 & 88.2\\\cmidrule(lr){2-3}\cmidrule(lr){4-5}\cmidrule(lr){6-7}\cmidrule(lr){8-9}\cmidrule(lr){10-11}
    \multirow{2}{*}{{\it morf.ww.pp.att}}
    & 92.3 & 91.9 & 91.6 & 89.3 & 89.0 & 89.6 & 70.9 & 72.0 & 92.2 & 91.1\\
    & 90.3 & 89.9 & 88.0 & 84.5 & 85.8 & 86.3 & 62.7 & 63.4 & 89.6 & 88.3\\\cmidrule(lr){2-3}\cmidrule(lr){4-5}\cmidrule(lr){6-7}\cmidrule(lr){8-9}\cmidrule(lr){10-11}
    \multirow{2}{*}{{\it morf.ww.mp.att}}
    & 92.2 & 91.9 & 91.3 & 89.0 & 88.9 & 90.0 & 70.9 & 71.8 & 92.3 & 91.0\\
    & 90.4 & 90.0 & 87.9 & 84.1 & 85.8 & 86.7 & 62.8 & 63.4 & 89.4 & 88.0\\\cmidrule(lr){2-3}\cmidrule(lr){4-5}\cmidrule(lr){6-7}\cmidrule(lr){8-9}\cmidrule(lr){10-11}
    \multirow{2}{*}{{\it morf.w-.p-.mtx}}
    & 92.1 & 91.9 & 91.4 & 88.9 & 89.3 & 89.8 & 70.6 & 71.9 & 92.2 & 90.6\\
    & 90.1 & 89.8 & 87.8 & 84.1 & 86.1 & 86.4 & 62.8 & 63.6 & 89.4 & 87.9\\\cmidrule(lr){2-3}\cmidrule(lr){4-5}\cmidrule(lr){6-7}\cmidrule(lr){8-9}\cmidrule(lr){10-11}
    \multirow{2}{*}{{\it morf.w-.pp.mtx}}
    & 92.4 & 91.8 & 91.3 & 88.8 & 89.3 & 90.0 & 71.0 & 72.2 & 92.3 & 91.3\\
    & 90.4 & 89.9 & 87.7 & 84.0 & 86.1 & 86.7 & 63.1 & 63.7 & 89.5 & 88.5\\\cmidrule(lr){2-3}\cmidrule(lr){4-5}\cmidrule(lr){6-7}\cmidrule(lr){8-9}\cmidrule(lr){10-11}
    \multirow{2}{*}{{\it morf.w-.mp.mtx}}
    & 92.2 & 92.1 & 91.4 & 88.8 & 89.2 & 90.3 & 70.2 & 72.2 & 92.2 & 90.5\\
    & 90.3 & 90.0 & 87.9 & 83.9 & 85.9 & 86.8 & 62.2 & 63.7 & 89.3 & 87.7\\\cmidrule(lr){2-3}\cmidrule(lr){4-5}\cmidrule(lr){6-7}\cmidrule(lr){8-9}\cmidrule(lr){10-11}
    \multirow{2}{*}{{\it morf.ww.p-.mtx}}
    & 92.2 & 91.8 & 91.3 & 88.9 & 89.3 & 89.6 & 70.1 & 71.4 & 92.5 & 91.4\\
    & 90.3 & 89.8 & 87.9 & 84.0 & 85.9 & 86.1 & 62.1 & 63.3 & 89.6 & 88.4\\\cmidrule(lr){2-3}\cmidrule(lr){4-5}\cmidrule(lr){6-7}\cmidrule(lr){8-9}\cmidrule(lr){10-11}
    \multirow{2}{*}{{\it morf.ww.pp.mtx}}
    & 92.2 & 91.7 & 91.4 & 89.0 & 89.2 & 89.9 & 71.3 & 72.1 & 92.6 & 90.8\\
    & 90.3 & 89.6 & 87.9 & 84.1 & 86.0 & 86.7 & 62.9 & 63.8 & 89.7 & 88.0\\\cmidrule(lr){2-3}\cmidrule(lr){4-5}\cmidrule(lr){6-7}\cmidrule(lr){8-9}\cmidrule(lr){10-11}
    \multirow{2}{*}{{\it morf.ww.mp.mtx}}
    & 92.3 & 91.8 & 91.2 & 89.1 & 89.5 & 89.8 & 71.1 & 72.2 & 92.1 & 90.8\\
    & 90.4 & 89.9 & 87.7 & 84.2 & 86.3 & 86.4 & 62.5 & 63.8 & 89.3 & 88.0\\\cmidrule(lr){2-3}\cmidrule(lr){4-5}\cmidrule(lr){6-7}\cmidrule(lr){8-9}\cmidrule(lr){10-11}
          
        \multirow{2}{*}{\texttt{sgns}}
        & 92.3 & 91.9 & 91.4 & 89.3 & 88.9 & 89.5 & 70.5 & 72.2 & 92.4 & 91.5\\
        & 90.4 & 89.8 & 87.9 & 84.4 & 85.6 & 86.2 & 62.5 & 63.5 & 89.8 & 88.7\\\cmidrule(lr){2-3}\cmidrule(lr){4-5}\cmidrule(lr){6-7}\cmidrule(lr){8-9}\cmidrule(lr){10-11}
        \multirow{2}{*}{\texttt{ft}}
        & 92.3 & 92.1 & 91.6 & 89.1 & 89.7 & 90.4 & 71.2 & 73.1 & 92.6 & 91.2\\
        & 90.3 & 90.2 & 87.9 & 84.4 & 86.9 & 87.1 & 63.3 & 65.1 & 89.7 & 88.3\\\cmidrule(lr){2-3}\cmidrule(lr){4-5}\cmidrule(lr){6-7}\cmidrule(lr){8-9}\cmidrule(lr){10-11}
        \multirow{2}{*}{\texttt{our\_ft}}
	& 92.7 & 92.2 & 91.6 & 89.6 & 90.5 & 90.9 & 71.5 & 73.0 & 92.52 & 91.6\\
	& 90.7 & 90.2 & 87.9 & 84.9 & 87.6 & 87.8 & 63.5 & 64.9 & 89.72 & 88.8\\
		\bottomrule       
    \end{tabularx}}
     \caption{Results on dependency parsing across languages for Morfessor ({\it morf}).} \label{tb:ps_morf} 
\end{table*}

\begin{table*}[t]
	\centering
	\def\arraystretch{0.85}
    {\footnotesize
	\begin{tabularx}{\textwidth}{l XXXXXXXXXX}
	    \toprule
	    & \textsc{dev(en)} & \textsc{test} 
	    & \textsc{dev(de)} & \textsc{test}
	    & \textsc{dev(fi)} & \textsc{test}
	    & \textsc{dev(tr)} & \textsc{test}
	    & \textsc{dev(he)} & \textsc{test}\\ \cmidrule(lr){2-3} \cmidrule(lr){4-5}\cmidrule(lr){6-7}\cmidrule(lr){8-9}\cmidrule(lr){10-11}
    \multirow{2}{*}{{\it bpe.w-.p-.add}}
    & 92.3 & 91.5 & 91.1 & 89.1 & 89.3 & 89.9 & 70.2 & 72.1 & 91.9 & 91.0\\
    & 90.3 & 89.6 & 87.6 & 84.2 & 85.9 & 86.4 & 62.6 & 63.9 & 89.2 & 88.2\\\cmidrule(lr){2-3}\cmidrule(lr){4-5}\cmidrule(lr){6-7}\cmidrule(lr){8-9}\cmidrule(lr){10-11}
    \multirow{2}{*}{{\it bpe.w-.pp.add}}
    & 92.2 & 91.6 & 91.5 & 89.1 & 89.0 & 90.0 & 71.2 & 71.6 & 92.0 & 90.9\\
    & 90.2 & 89.5 & 87.9 & 84.3 & 85.8 & 86.7 & 63.2 & 63.2 & 89.1 & 88.1\\\cmidrule(lr){2-3}\cmidrule(lr){4-5}\cmidrule(lr){6-7}\cmidrule(lr){8-9}\cmidrule(lr){10-11}
    \multirow{2}{*}{{\it bpe.w-.mp.add}}
    & 91.9 & 91.6 & 91.5 & 89.2 & 89.3 & 89.8 & 70.7 & 71.6 & 92.4 & 91.2\\
    & 90.0 & 89.7 & 87.9 & 84.3 & 86.1 & 86.3 & 62.7 & 63.4 & 89.7 & 88.2\\\cmidrule(lr){2-3}\cmidrule(lr){4-5}\cmidrule(lr){6-7}\cmidrule(lr){8-9}\cmidrule(lr){10-11}
    \multirow{2}{*}{{\it bpe.ww.p-.add}}
    & 92.1 & 91.8 & 91.3 & 89.2 & 89.5 & 89.8 & 70.2 & 71.4 & 92.5 & 91.1\\
    & 90.2 & 89.9 & 87.7 & 84.4 & 86.2 & 86.5 & 62.1 & 63.2 & 89.5 & 88.2\\\cmidrule(lr){2-3}\cmidrule(lr){4-5}\cmidrule(lr){6-7}\cmidrule(lr){8-9}\cmidrule(lr){10-11}
    \multirow{2}{*}{{\it bpe.ww.pp.add}}
    & 92.3 & 91.6 & 91.2 & 89.6 & 89.2 & 90.2 & 70.8 & 72.3 & 92.3 & 91.2\\
    & 90.4 & 89.6 & 87.7 & 84.7 & 86.2 & 87.0 & 62.9 & 64.2 & 89.4 & 88.4\\\cmidrule(lr){2-3}\cmidrule(lr){4-5}\cmidrule(lr){6-7}\cmidrule(lr){8-9}\cmidrule(lr){10-11}
    \multirow{2}{*}{{\it bpe.ww.mp.add}}
    & 92.3 & 91.9 & 91.6 & 89.2 & 89.9 & 90.7 & 71.2 & 72.3 & 92.2 & 91.0\\
    & 90.3 & 89.9 & 88.1 & 84.5 & 86.9 & 87.4 & 63.1 & 64.0 & 89.4 & 88.2\\\cmidrule(lr){2-3}\cmidrule(lr){4-5}\cmidrule(lr){6-7}\cmidrule(lr){8-9}\cmidrule(lr){10-11}
    \multirow{2}{*}{{\it bpe.w-.p-.att}}
    & 92.3 & 91.3 & 91.4 & 89.1 & 88.8 & 89.4 & 70.4 & 71.9 & 92.0 & 90.9\\
    & 90.3 & 89.4 & 87.9 & 84.2 & 85.3 & 85.7 & 62.3 & 63.0 & 89.1 & 87.9\\\cmidrule(lr){2-3}\cmidrule(lr){4-5}\cmidrule(lr){6-7}\cmidrule(lr){8-9}\cmidrule(lr){10-11}
    \multirow{2}{*}{{\it bpe.w-.pp.att}}
    & 92.1 & 91.6 & 91.4 & 88.6 & 88.7 & 89.5 & 70.9 & 71.5 & 91.9 & 91.0\\
    & 90.2 & 89.7 & 87.6 & 83.7 & 85.4 & 86.0 & 62.5 & 63.3 & 89.0 & 88.3\\\cmidrule(lr){2-3}\cmidrule(lr){4-5}\cmidrule(lr){6-7}\cmidrule(lr){8-9}\cmidrule(lr){10-11}
    \multirow{2}{*}{{\it bpe.w-.mp.att}}
    & 92.1 & 91.5 & 91.5 & 89.3 & 89.1 & 89.9 & 70.5 & 72.4 & 92.2 & 90.8\\
    & 90.2 & 89.6 & 87.9 & 84.4 & 85.8 & 86.3 & 62.5 & 63.9 & 89.5 & 88.0\\\cmidrule(lr){2-3}\cmidrule(lr){4-5}\cmidrule(lr){6-7}\cmidrule(lr){8-9}\cmidrule(lr){10-11}
    \multirow{2}{*}{{\it bpe.ww.p-.att}}
    & 92.3 & 92.0 & 91.4 & 88.6 & 89.1 & 90.1 & 70.8 & 71.3 & 92.0 & 90.8\\
    & 90.4 & 89.9 & 87.9 & 83.8 & 85.8 & 86.8 & 62.4 & 63.1 & 89.3 & 88.0\\\cmidrule(lr){2-3}\cmidrule(lr){4-5}\cmidrule(lr){6-7}\cmidrule(lr){8-9}\cmidrule(lr){10-11}
    \multirow{2}{*}{{\it bpe.ww.pp.att}}
    & 92.3 & 91.8 & 91.4 & 89.2 & 89.2 & 90.0 & 71.0 & 72.1 & 92.1 & 91.1\\
    & 90.5 & 89.8 & 87.7 & 84.6 & 85.9 & 86.4 & 63.2 & 64.0 & 89.5 & 88.2\\\cmidrule(lr){2-3}\cmidrule(lr){4-5}\cmidrule(lr){6-7}\cmidrule(lr){8-9}\cmidrule(lr){10-11}
    \multirow{2}{*}{{\it bpe.ww.mp.att}}
    & 92.5 & 92.0 & 91.5 & 89.3 & 89.5 & 90.3 & 70.6 & 71.7 & 91.9 & 90.9\\
    & 90.6 & 89.9 & 88.1 & 83.9 & 86.4 & 86.9 & 62.6 & 63.4 & 89.1 & 88.0\\\cmidrule(lr){2-3}\cmidrule(lr){4-5}\cmidrule(lr){6-7}\cmidrule(lr){8-9}\cmidrule(lr){10-11}
    \multirow{2}{*}{{\it bpe.w-.p-.mtx}}
    & 91.9 & 91.7 & 91.1 & 88.9 & 88.6 & 89.5 & 69.9 & 71.5 & 91.5 & 91.2\\
    & 90.1 & 89.7 & 87.5 & 83.9 & 85.2 & 85.9 & 61.9 & 63.1 & 88.7 & 88.3\\\cmidrule(lr){2-3}\cmidrule(lr){4-5}\cmidrule(lr){6-7}\cmidrule(lr){8-9}\cmidrule(lr){10-11}
    \multirow{2}{*}{{\it bpe.w-.pp.mtx}}
    & 92.1 & 91.7 & 91.1 & 88.8 & 89.0 & 89.5 & 70.7 & 71.8 & 92.0 & 91.0\\
    & 90.2 & 89.7 & 87.5 & 84.0 & 85.6 & 86.0 & 62.8 & 63.3 & 89.2 & 88.2\\\cmidrule(lr){2-3}\cmidrule(lr){4-5}\cmidrule(lr){6-7}\cmidrule(lr){8-9}\cmidrule(lr){10-11}
    \multirow{2}{*}{{\it bpe.w-.mp.mtx}}
    & 92.1 & 91.6 & 91.1 & 89.1 & 89.0 & 89.9 & 70.3 & 72.0 & 92.2 & 90.6\\
    & 90.2 & 89.5 & 87.6 & 84.1 & 85.7 & 86.3 & 62.6 & 63.6 & 89.3 & 87.7\\\cmidrule(lr){2-3}\cmidrule(lr){4-5}\cmidrule(lr){6-7}\cmidrule(lr){8-9}\cmidrule(lr){10-11}
    \multirow{2}{*}{{\it bpe.ww.p-.mtx}}
    & 92.3 & 92.1 & 91.6 & 89.0 & 89.3 & 90.0 & 71.7 & 72.3 & 91.9 & 90.5\\
    & 90.4 & 90.0 & 88.0 & 84.2 & 86.0 & 86.7 & 63.4 & 63.7 & 88.9 & 87.7\\\cmidrule(lr){2-3}\cmidrule(lr){4-5}\cmidrule(lr){6-7}\cmidrule(lr){8-9}\cmidrule(lr){10-11}
    \multirow{2}{*}{{\it bpe.ww.pp.mtx}}
    & 92.4 & 91.9 & 91.3 & 89.1 & 89.3 & 89.7 & 69.8 & 72.2 & 92.0 & 90.7\\
    & 90.4 & 89.8 & 87.7 & 84.2 & 85.9 & 86.2 & 62.0 & 63.8 & 89.2 & 87.9\\\cmidrule(lr){2-3}\cmidrule(lr){4-5}\cmidrule(lr){6-7}\cmidrule(lr){8-9}\cmidrule(lr){10-11}
    \multirow{2}{*}{{\it bpe.ww.mp.mtx}}
    & 92.3 & 91.7 & 91.2 & 89.4 & 89.8 & 90.2 & 71.2 & 72.3 & 92.2 & 91.0\\
    & 90.4 & 89.6 & 87.7 & 84.7 & 86.7 & 86.8 & 63.1 & 64.2 & 89.4 & 88.1\\\cmidrule(lr){2-3}\cmidrule(lr){4-5}\cmidrule(lr){6-7}\cmidrule(lr){8-9}\cmidrule(lr){10-11}
        
        \multirow{2}{*}{\texttt{sgns}}
        & 92.3 & 91.9 & 91.4 & 89.3 & 88.9 & 89.5 & 70.5 & 72.2 & 92.4 & 91.5\\
        & 90.4 & 89.8 & 87.9 & 84.4 & 85.6 & 86.2 & 62.5 & 63.5 & 89.8 & 88.7\\\cmidrule(lr){2-3}\cmidrule(lr){4-5}\cmidrule(lr){6-7}\cmidrule(lr){8-9}\cmidrule(lr){10-11}
        \multirow{2}{*}{\texttt{ft}}
        & 92.3 & 92.1 & 91.6 & 89.1 & 89.7 & 90.4 & 71.2 & 73.1 & 92.6 & 91.2\\
        & 90.3 & 90.2 & 87.9 & 84.4 & 86.9 & 87.1 & 63.3 & 65.1 & 89.7 & 88.3\\\cmidrule(lr){2-3}\cmidrule(lr){4-5}\cmidrule(lr){6-7}\cmidrule(lr){8-9}\cmidrule(lr){10-11}
        \multirow{2}{*}{\texttt{our\_ft}}
	& 92.7 & 92.2 & 91.6 & 89.6 & 90.5 & 90.9 & 71.5 & 73.0 & 92.52 & 91.6\\
	& 90.7 & 90.2 & 87.9 & 84.9 & 87.6 & 87.8 & 63.5 & 64.9 & 89.72 & 88.8\\
		\bottomrule       
    \end{tabularx}}
     \caption{Results on dependency parsing across languages for BPE ({\it bpe}).} \label{tb:ps_bpe} 
\end{table*}

\begin{table*}[t]
	\centering
	\def\arraystretch{0.85}
    {\footnotesize
	\begin{tabularx}{\textwidth}{l XXXXXXXXXX}
	    \toprule
	    & \textsc{dev(en)} & \textsc{test} 
	    & \textsc{dev(de)} & \textsc{test}
	    & \textsc{dev(fi)} & \textsc{test}
	    & \textsc{dev(tr)} & \textsc{test}
	    & \textsc{dev(he)} & \textsc{test}\\ \cmidrule(lr){2-3} \cmidrule(lr){4-5}\cmidrule(lr){6-7}\cmidrule(lr){8-9}\cmidrule(lr){10-11}
	    \multirow{1}{*}{{\it sms.w-.p-.add}}
& 52.88 & 52.57 & 51.09 & 50.95 & 55.15 & 55.08 & 53.87 & 53.52\\
\multirow{1}{*}{{\it sms.w-.st.add}}
& 52.68 & 52.7 & 51.2 & 51.23 & 55.38 & 55.13 & 54.03 & 53.68\\
\multirow{1}{*}{{\it sms.w-.pp.add}}
& 53.19 & 52.94 & 50.48 & 50.83 & 55.44 & 55.30 & 54.05 & 53.88\\
\multirow{1}{*}{{\it sms.w-.mp.add}}
& 52.87 & 52.76 & 51.1 & 51.11 & 55.18 & 55.00 & 54.24 & 53.96\\
\multirow{1}{*}{{\it sms.ww.p-.add}}
& 53.56 & 53.69 & 51.73 & 51.77 & 55.92 & 55.84 & 55.01 & 54.76\\
\multirow{1}{*}{{\it sms.ww.st.add}}
& 54.21 & 54.26 & 52.53 & 52.63 & 56.21 & 55.95 & 55.96 & 55.43\\
\multirow{1}{*}{{\it sms.ww.pp.add}}
& 54.51 & 54.66 & 52.98 & 53.04 & 56.06 & 56.00 & 55.92 & 55.75\\
\multirow{1}{*}{{\it sms.ww.mp.add}}
& 54.27 & 54.43 & 52.77 & 52.88 & 56.04 & 55.71 & 56.38 & 55.99\\
\multirow{1}{*}{{\it sms.w-.p-.att}}
& 52.81 & 52.7 & 50.7 & 50.8 & 54.85 & 54.67 & 53.62 & 53.45\\
\multirow{1}{*}{{\it sms.w-.st.att}}
& 52.57 & 52.64 & 51.64 & 51.65 & 54.87 & 54.69 & 53.27 & 53.14\\
\multirow{1}{*}{{\it sms.w-.pp.att}}
& 52.94 & 52.66 & 52.1 & 52.5 & 54.91 & 54.83 & 53.39 & 53.21\\
\multirow{1}{*}{{\it sms.w-.mp.att}}
& 52.99 & 52.91 & 51.7 & 51.9 & 55.02 & 55.03 & 54.06 & 53.72\\
\multirow{1}{*}{{\it sms.ww.p-.att}}
& 53.68 & 53.67 & 51.48 & 51.69 & 55.59 & 55.73 & 55.17 & 54.92\\
\multirow{1}{*}{{\it sms.ww.st.att}}
& 53.55 & 53.66 & 51.64 & 51.65 & 55.27 & 55.35 & 53.84 & 53.26\\
\multirow{1}{*}{{\it sms.ww.pp.att}}
& 53.14 & 53.08 & 52.1 & 52.5 & 54.66 & 54.52 & 54.95 & 54.77\\
\multirow{1}{*}{{\it sms.ww.mp.att}}
& 53.82 & 53.79 & 51.67 & 51.89 & 54.02 & 54.04 & 54.1 & 53.76\\
\multirow{1}{*}{{\it sms.w-.p-.mtx}}
& 52.7 & 52.67 & 50.29 & 50.52 & 54.61 & 54.51 & 53.44 & 53.24\\
\multirow{1}{*}{{\it sms.w-.st.mtx}}
& 52.75 & 52.61 & 50.42 & 50.85 & 54.15 & 54.09 & 53.44 & 52.84\\
\multirow{1}{*}{{\it sms.w-.pp.mtx}}
& 52.89 & 52.87 & 51.07 & 51.38 & 54.58 & 54.70 & 53.42 & 53.16\\
\multirow{1}{*}{{\it sms.w-.mp.mtx}}
& 53.03 & 52.89 & 50.52 & 50.95 & 55.19 & 54.90 & 54.23 & 53.68\\
\multirow{1}{*}{{\it sms.ww.p-.mtx}}
& 53.75 & 53.81 & 51.47 & 51.48 & 55.34 & 55.37 & 54.51 & 54.36\\
\multirow{1}{*}{{\it sms.ww.st.mtx}}
& 53.75 & 53.64 & 51.24 & 51.51 & 55.4 & 55.36 & 54.97 & 54.72\\
\multirow{1}{*}{{\it sms.ww.pp.mtx}}
& 53.80 & 53.96 & 51.92 & 52.1 & 54.57 & 54.54 & 54.94 & 54.57\\
\multirow{1}{*}{{\it sms.ww.mp.mtx}}
& 54.23 & 54.2 & 52 & 52.2 & 55.02 & 54.84 & 55.8 & 55.45\\
\multirow{1}{*}{{\it morf.w-.p-.add}}
& 54.86 & 55.35 & 53.41 & 53.28 & 57.01 & 56.94 & 55.92 & 55.8 & 60.37 & 59.67\\
\multirow{1}{*}{{\it morf.w-.pp.add}}
& 55.55 & 55.57 & 54.02 & 54.06 & 57.69 & 57.41 & 56.79 & 56.31 & 60.61 & 59.99\\
\multirow{1}{*}{{\it morf.w-.mp.add}}
& 55.01 & 55.18 & 53.5 & 53.35 & 57.09 & 56.82 & 55.26 & 54.7 & 60.1 & 59.24\\
\multirow{1}{*}{{\it morf.ww.p-.add}}
& 54.94 & 55.09 & 53.27 & 53.02 & 56.61 & 56.55 & 55.5 & 54.9 & 60.23 & 59.65\\
\multirow{1}{*}{{\it morf.ww.pp.add}}
& 55.20 & 55.60 & 53.83 & 53.82 & 57.32 & 57.2 & 56.41 & 55.92 & 60.57 & 60.09\\
\multirow{1}{*}{{\it morf.ww.mp.add}}
& 54.63 & 54.67 & 53.46 & 53.4 & 56.81 & 56.73 & 54.42 & 53.88 & 59.67 & 58.87\\
\multirow{1}{*}{{\it morf.w-.p-.att}}
& 54.84 & 55.02 & 52.98 & 52.91 & 56.93 & 56.83 & 55.86 & 55.52 & 60.3 & 59.78\\
\multirow{1}{*}{{\it morf.w-.pp.att}}
& 55.28 & 55.68 & 53.92 & 54.01 & 57.36 & 57.24 & 56.62 & 56.13 & 60.83 & 60.34\\
\multirow{1}{*}{{\it morf.w-.mp.att}}
& 55.4 & 55.3 & 53.72 & 53.7 & 56.85 & 56.93 & 55.24 & 54.6 & 59.89 & 59.38\\
\multirow{1}{*}{{\it morf.ww.p-.att}}
& 54.94 & 54.87 & 53.07 & 53.12 & 56.7 & 56.36 & 55.5 & 55.07 & 60.31 & 59.73\\
\multirow{1}{*}{{\it morf.ww.pp.att}}
& 55.26 & 55.35 & 53.83 & 53.92 & 57.39 & 57.15 & 56.39 & 55.99 & 60.5 & 59.94\\
\multirow{1}{*}{{\it morf.ww.mp.att}}
& 55.16 & 55.27 & 53.22 & 53.26 & 56.84 & 56.67 & 54.89 & 54.32 & 59.59 & 59.04\\
\multirow{1}{*}{{\it morf.w-.p-.mtx}}
& 54.97 & 55.06 & 52.39 & 52.42 & 56.76 & 56.88 & 55.93 & 55.39 & 60.26 & 59.66\\
\multirow{1}{*}{{\it morf.w-.pp.mtx}}
& 55.18 & 55.4 & 53.95 & 53.92 & 57.49 & 57.38 & 56.62 & 56.14 & 60.65 & 60.02\\
\multirow{1}{*}{{\it morf.w-.mp.mtx}}
& 55.18 & 55.25 & 52.92 & 53.01 & 56.97 & 56.69 & 55.02 & 54.48 & 60.07 & 59.36\\
\multirow{1}{*}{{\it morf.ww.p-.mtx}}
& 54.9 & 55.07 & 53.06 & 53.07 & 56.61 & 56.63 & 55.6 & 54.87 & 60.3 & 59.65\\
\multirow{1}{*}{{\it morf.ww.pp.mtx}}
& 55.25 & 55.45 & 53.64 & 53.55 & 57.31 & 57.17 & 56.28 & 55.86 & 60.52 & 59.98\\
\multirow{1}{*}{{\it morf.ww.mp.mtx}}
& 55.1 & 55.16 & 53.43 & 53.4 & 56.81 & 56.69 & 54.57 & 53.89 & 59.71 & 58.95\\
\multirow{1}{*}{{\it bpe.w-.p-.add}}
& 50.74 & 51.3 & 50.43 & 50.42 & 52.74 & 52.53 & 47.85 & 47.6 & 51.67 & 51.44\\
\multirow{1}{*}{{\it bpe.w-.pp.add}}
& 50.92 & 51.59 & 50.77 & 50.68 & 52.93 & 52.74 & 48.32 & 47.71 & 51.65 & 51.33\\
\multirow{1}{*}{{\it bpe.w-.mp.add}}
& 51.35 & 51.93 & 51 & 50.81 & 53.37 & 53.1 & 47.71 & 47.3 & 52.07 & 51.78\\
\multirow{1}{*}{{\it bpe.ww.p-.add}}
& 54.6 & 55.01 & 53.45 & 53.39 & 56.38 & 55.96 & 53.35 & 53.05 & 59.53 & 59.03\\
\multirow{1}{*}{{\it bpe.ww.pp.add}}
& 54.86 & 55.03 & 53.47 & 53.53 & 56.65 & 56.27 & 53.03 & 52.71 & 59.95 & 59.26\\
\multirow{1}{*}{{\it bpe.ww.mp.add}}
& 55.28 & 55.7 & 53.90 & 53.61 & 57.14 & 56.82 & 53.9 & 53.52 & 60.1 & 59.24\\
\multirow{1}{*}{{\it bpe.w-.p-.att}}
& 50.65 & 51.15 & 50.35 & 50.21 & 52.71 & 52.18 & 47.52 & 47.05 & 51.45 & 51.31\\
\multirow{1}{*}{{\it bpe.w-.pp.att}}
& 50.85 & 51.57 & 50.66 & 50.45 & 52.94 & 52.47 & 47.94 & 47.58 & 51.71 & 51.35\\
\multirow{1}{*}{{\it bpe.w-.mp.att}}
& 51.28 & 51.94 & 50.92 & 50.89 & 53.13 & 52.93 & 47.51 & 47.21 & 52.05 & 51.69\\
\multirow{1}{*}{{\it bpe.ww.p-.att}}
& 53.19 & 53.49 & 51.99 & 52.09 & 54.96 & 54.71 & 52.3 & 52.37 & 57.6 & 57.17\\
\multirow{1}{*}{{\it bpe.ww.pp.att}}
& 53.18 & 53.56 & 52.13 & 51.92 & 55.13 & 54.91 & 53.41 & 53.29 & 58.67 & 57.96\\
\multirow{1}{*}{{\it bpe.ww.mp.att}}
& 54.23 & 54.71 & 52.88 & 52.78 & 55.85 & 55.64 & 53.02 & 53.15 & 58.87 & 58.21\\
\multirow{1}{*}{{\it bpe.w-.p-.mtx}}
& 50.81 & 51.48 & 50.42 & 50.33 & 52.76 & 52.36 & 47.58 & 47.14 & 51.67 & 51.34\\
\multirow{1}{*}{{\it bpe.w-.pp.mtx}}
& 51.05 & 51.79 & 50.53 & 50.4 & 52.89 & 52.29 & 47.36 & 46.97 & 51.79 & 51.46\\
\multirow{1}{*}{{\it bpe.w-.mp.mtx}}
& 51.45 & 51.88 & 51.15 & 50.93 & 53.27 & 52.93 & 47.7 & 47.34 & 52.18 & 51.87\\
\multirow{1}{*}{{\it bpe.ww.p-.mtx}}
& 53.19 & 53.49 & 52.35 & 52.07 & 55.15 & 54.77 & 52.87 & 52.6 & 58.43 & 57.84\\
\multirow{1}{*}{{\it bpe.ww.pp.mtx}}
& 53.17 & 53.57 & 52.35 & 52.18 & 55.08 & 54.58 & 52.44 & 52.02 & 57.92 & 57.43\\
\multirow{1}{*}{{\it bpe.ww.mp.mtx}}
& 54.64 & 55.05 & 53.11 & 52.86 & 55.73 & 55.49 & 53.34 & 53.11 & 59.16 & 58.37\\
\multirow{1}{*}{\texttt{sgns}}
& 50.56 & 51 & 50.55 & 50.14 & 49.88 & 49.87 & 55.09 & 54.35 & 54.93 & 54.55\\
\multirow{1}{*}{\texttt{ft}}
& 54.97 & 55.15 & 54.46 & 54.55 & 57.18 & 57.18 & 55.46 & 54.62 & 59.67 & 59.09\\
\multirow{1}{*}{\texttt{our\_ft}}
& 54.59 & 55.12 & 54.26 & 54.4 & 57.67 & 57.57 & 56.19 & 55.73 & 60.05 & 59.65\\
		\bottomrule       
    \end{tabularx}}
     \caption{Accuracy on fine-grained entity typing across languages.} \label{tb:et2} 
\end{table*}

\end{document}